%% file: main.tex
\documentclass[pmlr]{jmlr_arxiv}
\makeatletter
\let\Ginclude@graphics\@org@Ginclude@graphics
\makeatother
\jmlrworkshop{\href{https://www.mlforhc.org}{Machine Learning for Healthcare, 2021}}
\jmlryear{Last updated: 2021-07-21}
\jmlrvolume{}
\jmlrpages{}

\usepackage{url}
\usepackage{hyperref}
\hypersetup{
    colorlinks,
    citecolor=blue,
    filecolor=blue,
    linkcolor=blue,
    urlcolor=blue,
}

\usepackage{xcolor}

\setcitestyle{authoryear,open={(},close={)}}

\usepackage[utf8]{inputenc} 
\usepackage[T1]{fontenc}    
\usepackage{hyperref}       
\usepackage{url}            
\usepackage{microtype}      
\usepackage{nicefrac}       
\usepackage{enumitem}

\usepackage{graphicx}
\usepackage{caption}
\usepackage{chngcntr} 
\usepackage{multirow}
\usepackage{booktabs}
\usepackage{rotating}
\usepackage{array}
\usepackage{float}
\newcolumntype{P}[1]{>{\centering\arraybackslash}p{#1}}

\DeclareMathOperator*{\argmax}{arg\,max}

\newcommand{\KL}[0]{\text{KL}}

\newcommand{\datasetName}[0]{TMED}
\newcommand{\datasetURL}[0]{\url{https://\datasetName.cs.tufts.edu}}
\newcommand{\codeURL}[0]{\url{https://github.com/tufts-ml/ssl-for-echocardiograms}}

\begin{document}

\setlength{\abovedisplayskip}{2pt plus 3pt}
\setlength{\belowdisplayskip}{2pt plus 3pt}

\title[SSL for Classifying View and Diagnosis Aortic Stenosis from Echocardiograms]{A New Semi-supervised Learning Benchmark for Classifying View and Diagnosing Aortic Stenosis from Echocardiograms}

\author{\Name{Zhe Huang}$^{1}$, M.S.
		\Email{\textsc{zhe.huang@tufts.edu}}
\AND
        \Name{Gary Long}$^{2}$, M.S.
		\Email{\textsc{gary@cvai.solutions}}
\AND
        \Name{Benjamin Wessler}$^{3}$, M.D.
		\Email{\textsc{bwessler@tuftsmedicalcenter.org}}
\AND       
        \Name{Michael C. Hughes}$^1$, Ph.D.
        \Email{\textsc{michael.hughes@tufts.edu}}
        \\
        \addr $^1$ Dept. of Computer Science, Tufts University, Medford, MA, USA
        \\
        \addr $^2$ CVAI Solutions, Dorchester, MA        \\
        \addr $^3$ Division of Cardiology, Tufts Medical Center, Boston, MA
}

\maketitle

\begin{abstract}
  \input{abstract.tex}
\end{abstract}

\section{Introduction}
\label{sec:Introduction}
\input{intro.tex}

\section{Background and Related Work}
\label{sec:RelatedWork}
\input{bg_and_related_work.tex}

\input{fig_view_samples.tex}

\section{Dataset}
\label{sec:Dataset}
\input{dataset.tex}

\section{Methods}
\label{sec:Methods}

\input{methods.tex}


\section{Results}
\label{sec:Results}
\input{results.tex}

\input{fig_confusion_mat.tex}

\section{Discussion}
\label{sec:Discussion}
\input{discussion.tex}

\input{acks.tex}

\bibliography{ssl_for_echo.bib}

\appendix

\makeatletter
\let\c@table\c@figure 
\let\ftype@table\ftype@figure 
\makeatother

\counterwithin{table}{section}
\setcounter{table}{0}
\counterwithin{figure}{section}
\setcounter{figure}{0}
\counterwithin{algocf}{section} 
\setcounter{algocf}{0}

\input{Appendix.tex}

\end{document}

%% file: abstract.tex
Semi-supervised image classification has shown substantial progress in learning from limited labeled data, but recent advances remain largely untested for clinical applications.
Motivated by the urgent need to improve timely diagnosis of life-threatening heart conditions, especially aortic stenosis,
we develop a benchmark dataset to assess semi-supervised approaches to two tasks relevant to cardiac ultrasound (echocardiogram) interpretation: view classification and disease severity classification.
We find that a state-of-the-art method called MixMatch achieves promising gains in heldout accuracy on both tasks, learning from a large volume of truly unlabeled images as well as a labeled set collected at great expense to achieve better performance than is possible with the labeled set alone.
We further pursue \emph{patient-level} diagnosis prediction, which requires aggregating across hundreds of images of diverse view types, most of which are irrelevant, to make a coherent prediction.
The best patient-level performance is achieved by new methods that \emph{prioritize} diagnosis predictions from images that are predicted to be clinically-relevant views and \emph{transfer} knowledge from the view task to the diagnosis task.
We hope our released dataset\footnote{Tufts Medical Echocardiogram Dataset (\datasetName): \datasetURL} and evaluation framework\footnote{Open-source code: \codeURL} inspire further improvements in multi-task semi-supervised learning for clinical applications.

%% file: intro.tex
Our motivating task is to improve timely diagnosis and treatment of aortic stenosis (AS), a common cardiac valve condition.
If left untreated, severe AS has lower 5-year survival rates than several metastatic cancers~\citep{howladerSEERCancerStatistics2020,clarkFiveyearClinicalEconomic2012}.
With timely diagnosis and surgical or transcatheter aortic valve replacement, AS becomes a treatable condition with very low mortality~\citep{lancellottiOutcomesPatientsAsymptomatic2018}.
Unfortunately, in current practice up to 2/3 of symptomatic AS patients may never get referred for care~\citep{tangContemporaryReasonsClinical2018,brennanProviderlevelVariabilityTreatment2019}.
There is an urgent need in improve timely detection of this life-threatening condition.
In this study, we develop and validate machine learning methods for automating the preliminary interpretation of cardiac ultrasound (echocardiogram) images, with the goal of expanding access to rapid and accurate diagnosis of AS while overcoming constraints on the availability of labeled data needed to train methods effectively.

Recent advances in computer vision and machine learning have made it possible to \emph{automate} the way medical images are turned into actionable knowledge for diagnosing and treating disease across cardiology~\citep{chenDeepLearningCardiac2020}, radiology, and other areas of medicine~\citep{shenDeepLearningMedical2017}.
However, in order to work well, modern deep learning methods require \emph{large} amounts of \emph{labeled} training examples, where each labeled example consists of an image and its desired class label.
While images themselves are routinely collected and easily available in electronic health records, obtaining \emph{labels} for images often requires manual effort from a clinical expert.
Thus, a key barrier to deploying deep learning image classifiers for specialty areas of medicine is that it is prohibitively \emph{difficult} and \emph{expensive} to acquire a large labeled dataset whose scale matches the tens of thousands of labeled examples available in common non-medical benchmarks such as CIFAR-10~\citep{krizhevskyLearningMultipleLayers2009} or Street View Housing Numbers~(SVHN,~\citet{netzerReadingDigitsNatural2011}). Privacy and regulatory concerns further inhibit sharing of labeled datasets even if they are collected within a single healthcare system.


A promising technology to overcome the need for abundant labeled data is semi-supervised learning (SSL)~\citep{zhuSemiSupervisedLearningLiterature2005,chapelleSemiSupervisedLearning2010,vanengelenSurveySemisupervisedLearning2020}.
SSL methods train classifiers simultaneously from two data sources: a \emph{small} labeled set and a \emph{large} unlabeled set.
Recent semi-supervised deep learning methods have produced remarkable progress~\citep{miyatoVirtualAdversarialTraining2019,berthelotRemixmatchSemisupervisedLearning2019,berthelotMixmatchHolisticApproach2019,sohnFixmatchSimplifyingSemisupervised2020,xie2019unsupervised,chenBigSelfSupervisedModels2020}.
On the standard SVHN image classification benchmark, a typical WideResNet achieves an error rate of 12.8\% when trained using only a \emph{small} labeled set of 1000 total labeled images (100 for each of 10 digit categories).
In contrast, a recent SSL method called MixMatch~\citep{berthelotMixmatchHolisticApproach2019} can reduce error rate to 3.3\% using the small labeled set plus 60,000 unlabeled examples, or even to 2.2\% with 600,000 unlabeled examples.
These improvements are comparable to training on 50x larger \emph{fully-labeled} datasets, but avoid the time and expense of collecting so many labels (which for our intended applications are reliably obtained only from clinical experts).
While SSL methods are promising, the application of modern SSL methods to real medical imaging tasks is largely \emph{untested} and requires development of new methods to address issues such as class imbalance and the need to aggregate many image-specific predictions into coherent decisions for the whole patient. 

In parallel to our work, several recent efforts do explore modern SSL methods to analyze medical images. \citet{meng2020mutual} proposed a SSL domain adaptation method for classifying views of fetal ultrasounds. \citet{calderon2021dealing} utilize MixMatch to detect COVID-19 based on chest X-ray images. \citet{chen2021venibot} leverage unlabeled data to improve vein segmentation performance using ultrasound images. \citet{wang2021deep} propose an SSL method for classifying breast lesion and ophthalmic diseases.
Our study contributes to this growing literature by applying SSL to improve patient-level screening in cardiology.

As a case study to test the promise of SSL methods, we develop SSL classifiers to diagnose aortic stenosis (AS). AS is diagnosed using ultrasound imaging of the heart, known as \emph{echocardiography}.
While ultrasound imaging itself is widely available and performed routinely for many patients for a variety of reasons, accurate interpretation of echocardiograms to make complex imaging diagnoses such as AS requires significant expertise that is not widely available.
Diagnostic errors may contribute to treatment delays because assessment is challenging and requires integrating information across many hemodynamic parameters~\citep{baumgartnerRecommendationsEchocardiographicAssessment2017}, that are often discordant~\citep{minnersInconsistenciesEchocardiographicCriteria2008} and have low inter-reader reliability~\citep{sacchiDopplerAssessmentAortic2018}.
Automated grading of AS has the potential to increase the accuracy and reproducibility of disease detection and reduce barriers to access~\citep{batchelorAorticValveStenosis2019}, especially as a first-line screening tool in geographic areas without expert cardiologists.
We believe that automated preliminary assessment of AS, with timely follow-up care by an expert clinical team, will improve patient outcomes by better identifying patients with this life-threatening condition that require treatment.


\begin{figure}[!t]
	\centering
	\includegraphics[width=0.9\textwidth]{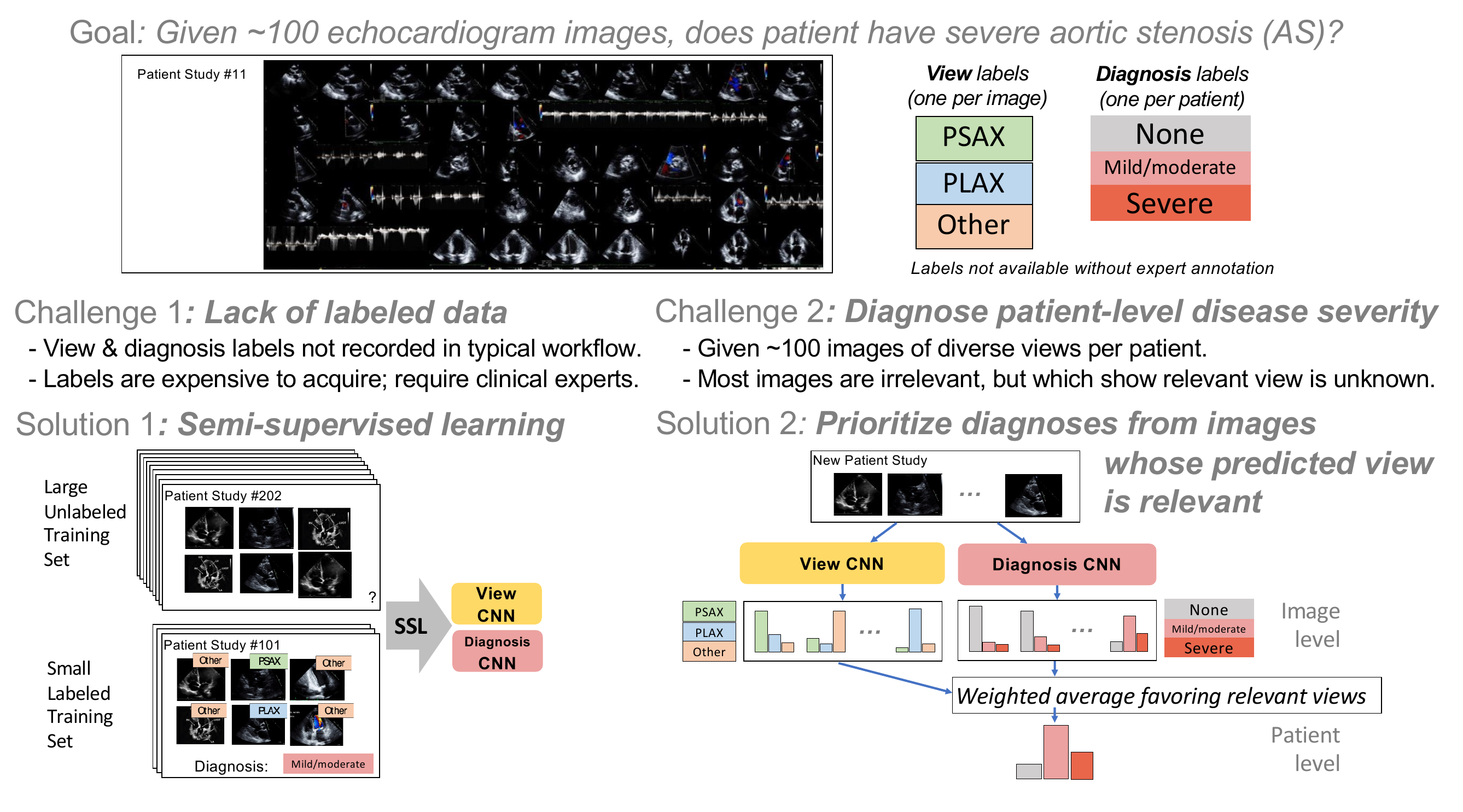}
    \caption{Illustration of our study's goal -- automating diagnosis of the severity of \emph{aortic stenosis} (AS) given hundreds of echocardiogram images collected in a typical exam -- as well as the key technical challenges and proposed contributions that address these challenges.}
    \label{fig:ssl_for_echo_diagram}
\end{figure}

\subsection*{Contributions}
In this work, we make the following contributions:
\begin{enumerate}[leftmargin=0cm,itemindent=.4cm,labelwidth=\itemindent,labelsep=0cm,align=left]
	\item 
	\textbf{New open-access SSL dataset -- the Tufts Medical Echocardiogram Dataset (TMED) -- to benchmark view and disease severity classification.}
	Our dataset is directly inspired by the need for automated preliminary assessment of aortic stenosis (AS).
	The labeled set of 260 patients contains an AS disease severity label for each patient as well as a view label for all images, all provided by expert clinicians.
	Furthermore, our dataset is designed to assess the true potential of semi-supervised learning, because in addition to the labeled set it contains a large \emph{unlabeled} set from 2645 patients captured in the course of standard cardiac care.
	Common SSL benchmarks such as CIFAR-10~\citep{krizhevskyLearningMultipleLayers2009} or STL-10~\citep{coatesAnalysisSingleLayer2011} do not contain truly unlabeled data but instead artificially ``forget'' existing labels.
	This makes the unlabeled data used in these benchmarks unrealistically clean, class-balanced, and relevant to the task.
	We hope our new dataset will inspire work on effectively using minimally-curated unlabeled data to improve medical image understanding.
	
	\item 
	\textbf{Evaluation of SSL methodology to find what works best and why.} We carefully compare standardized implementations of several state-of-the-art SSL methods. We find that MixMatch~\citep{berthelotMixmatchHolisticApproach2019} performs best on both view and diagnosis tasks, reliably beating labeled-set-only methods by over 2\% balanced accuracy on view classification (see Table~\ref{tab:view classification large}) and by over 3\% balanced accuracy on patient-level diagnosis (see Table~\ref{tab:diagnosis classification large patient}). These gains are also consistent for smaller versions of our dataset.
	  	Further ablation studies suggest that the surprisingly effective ``mix-up'' data augmentation strategy underlying MixMatch is a primary reason for success.
	  	In contrast, virtual adversarial training (VAT,~\citet{miyatoVirtualAdversarialTraining2019}), which by its adversarial design might be expected to perform better in medical imaging domains, only marginally improves over labeled-set-only methods.
	  
	\item \textbf{New methods for patient-level severity diagnosis without manually preselecting relevant views.}
	A patient study may contain over 100 echocardiogram images from diverse view types (see Fig.~\ref{fig:ssl_for_echo_diagram}).
	Only some of these images are relevant to diagnosing AS (only some views show the aortic valve that AS impacts).
	View type labels are not usually available.
	Previous work has relied on manual preselection of relevant views~\citep{madaniDeepEchocardiographyDataefficient2018,ouyangVideobasedAIBeattobeat2020}.
	We develop methods that directly consume \emph{all available images}, as would be needed in a fully automated deployment.	
	Rather than simply average diagnostic predictions across all images, we \emph{prioritize} diagnoses from images that our view classifiers suggest belong to the view types known to be relevant for AS.
	Using only the labeled set and trying to distinguish between 3 levels of AS severity (none, mild/moderate, and severe), prioritizing relevant views achieves patient-level balanced accuracy of 86.6\% compared to 81.6\% with simple averaging and 33\% for random guessing.
	Our best SSL methods that prioritize relevant views achieve 87.9\%, further improving to 90.1\% when we \emph{pretrain} a view classifier and use this to warm start our diagnosis classifier.
	These results suggest that fully-automated preliminary diagnosis of aortic stenosis may be soon realizable in practice.

\end{enumerate}

\subsection*{Generalizable Insights about Machine Learning in the Context of Healthcare}

The key barrier to applying deep learning to address many medical tasks is the lack of abundant labeled data.
Recent SSL methods appear to reach competitive performance with only modest labeled datasets, but lack authentic evaluation using truly unlabeled clinical data exhibiting issues such as irrelevance to the task or class imbalance.
Thus, existing benchmarks may be too ``clean'' and over-state the potential of SSL methods.
As a remedy, this study develops clinically-motivated semi-supervised learning tasks and releases a dataset useful for measuring the gains possible with SSL methods.
Next, we offer insight about \emph{which SSL methods work best and why}.
Our careful evaluation and ablation studies can help practitioners identify promising methods like MixMatch.
Furthermore, our analysis encourages work to \emph{look beyond simple averaging} when aggregating across multiple images to make a patient-level diagnosis.
Overall, we hope our work helps unlock the potential of easily-accessible unlabeled data to improve patient outcomes.

%% file: bg_and_related_work.tex
\subsection{Background: Single-task SSL with Neural Networks for Image Classification}
\label{sec:ssl_background} 

We consider the problem of \emph{semi-supervised} image classification.
For training, we are given two datasets.
First, a (small) \emph{labeled} dataset $\mathcal{D}^L$ containing $N^L$ pairs of images $x$ and corresponding labels $y$.
Second, a (large) \emph{unlabeled} dataset $\mathcal{D}^U$ of $N^U$ examples of images only, presumed to be sampled from a similar distribution as the labeled set.
Each image $x$ is represented as a standard tensor of pixel intensity values (one entry for each pixel and each color channel).
Each image-specific label $y \in \mathcal{C}$ indicates one of the possible classes in set $\mathcal{C}$.

Given a deep neural network parameterized by weight vector $\theta$, a standard SSL training procedure tries to find an (approximate) solution to the following loss minimization problem:
\begin{align}
\theta^* = \arg\min_{\theta}
	\frac{1}{N^L}
	\sum_{x, y \in \mathcal{D}^L}
	\mathcal{L}^L(y, x, \theta) 
	+\lambda 
	\frac{1}{N^U}
	\sum_{x \in \mathcal{D}^U} \mathcal{L}^U( x, \theta )
\end{align}
\label{eq:standard-SSL-loss-template}
Here, $\mathcal{L}^L$ indicates a \emph{labeled loss} function such as cross-entropy, and $\mathcal{L}^U$ indicates an \emph{unlabeled loss}. We show how each SSL method we explore instantiates this framework in Sec.~\ref{sec:Methods-SSL}.

For our specific application, we separately train neural networks to perform two tasks.
First, for \emph{view} classification we wish to map each image $x$ to a real-valued score for each of the possible view classes in set $\mathcal{C}_V$.
We use a network $f$ parameterized by weights $\theta_V$.
To obtain probability distributions over the classes, we use the softmax transformation, denoted as $S(\cdot)$, which produces a probability vector with same size as the given input vector.
Thus, the probability that an image's view label indicates the $c$-th element of $\mathcal{C}_V$ is then $S( f_{\theta_V}(x) )_c$.

Second, for \emph{diagnosis} classification we wish to map each image $x$ to a real-valued vector containing scores for each of 3 posssible diagnoses (no AS, mild/moderate AS, severe AS) in set $\mathcal{C}_D$.
We use network $g$ with weights $\theta_D$. We always use the same architecture as view network $f$.
The probability of assigning the $c$-th label in $\mathcal{C}_D$ to image $x$ is $S( g_{\theta_D}(x) )_c$.

\subsection{Semi-supervised learning: Datasets and evaluation}

Our work builds upon a previous study by~\citet{oliverRealisticEvaluationDeep2018} on best practices for SSL evaluation.
That work emphasizes the need to compare SSL methods that claim advantages from unlabeled data to strong baselines that only use the labeled set.
We follow this best practice in our work.
Their independent evaluation suggests that, if trained properly, modern deep SSL methods can leverage large unlabeled sets to obtain meaningful improvements.

However, a key limitation of existing SSL research is that popular evaluations, even those used in~\citet{oliverRealisticEvaluationDeep2018}, are confined to well-worn non-medical datasets, such as SVHN~\citep{netzerReadingDigitsNatural2011} or CIFAR-10~\citep{krizhevskyLearningMultipleLayers2009}.
These datasets consist of carefully curated \emph{balanced} class distributions, where the so-called ``unlabeled'' set is obtained by forgetting known labels.
Performance numbers are likely too optimistic compared to using truly unlabeled images.
As performance saturates on these benchmarks,
 SSL research needs to move towards datasets like ours featuring truly unlabeled images.
 
\subsection{Related Work: Supervised learning for cardiology}

Deep learning research efforts for cardiac imaging applications are plentiful~\citep{chenDeepLearningCardiac2020}.
For echocardiogram analysis, the task of \emph{view} classification has been pursued by several efforts~\citep{madaniFastAccurateView2018,zhangFullyAutomatedEchocardiogram2018,longIdentificationEchocardiographicImaging2018}.
Accuracies above 90\% can be achieved given a \emph{large labeled} dataset of echocardiogram images and corresponding views.
Training CNN view classifiers on 200,000 images from 240 patients,~\citet{madaniFastAccurateView2018} report accuracy of 91.7\% on 15 view types using low-resolution images, exceeding the performance of board-certified cardiographers.
\citet{zhangFullyAutomatedEchocardiogram2018} also report promising performance for CNNs to classify 23 different view types (e.g. 96\% accuracy for the PLAX view).
However, we emphasize that these efforts use \emph{proprietary datasets} that other researchers cannot build upon and extend. They also require \emph{large} labeled sets and do not address the key motivation of our work, the downstream diagnosis of \emph{aortic stenosis}.

Recently, researchers at Stanford developed an ``EchoNet'' deep learning methodology for echocardiography~\citep{ghorbaniDeepLearningInterpretation2020}, predicting measurements related to ejection fractions as well as some binary decisions such as ``does the patient have a pace-maker?'' or ``is there left ventricular hypertrophy?''
\citet{ghorbaniDeepLearningInterpretation2020} produce \emph{patient-level} decisions just by simple averaging, reporting the surprising conclusion that ``alternative methods were explored in order to aggregate frame-level predictions into one patient-level prediction and did not yield better results compared to simple averaging.''
Our later results suggest that smarter aggregation \emph{does} produce notable benefits.

The same team later produced ``EchoNet Dynamic'' methodology to learn from videos (not static images)~
\citep{ouyangVideobasedAIBeattobeat2020}.
releasing 10,030 videos of echocardiogram imagery for one particular view (apical-4 chamber) and associated measurement and diagnostic labels for thousands of subjects.
This is a welcome step forward, but this dataset pursues different goals than ours: there is no focus on semi-supervised learning and this data focuses on only one view type (apical 4 chamber or A4C) out of the dozens of possible views that make up a complete study.
While this A4C view is relevant to other measurement and diagnostic tasks in cardiology, it is not helpful for assessing valvular heart diseases generally nor AS in particular.

Another public echocardiogram dataset is the CAMUS dataset \citep{leclerc2019deep}. CAMUS' purpose is to evaluate \emph{segmentation} methods. The data contains images of apical 2 chamber (A2C) and apical 4 chamber (A4C) views from 500 patients, together with detailed annotations of anatomical structures (e.g. the myocardium and the left atrium).  Like other prior work, these views are not relevant for the diagnosis of aortic valve disease.

Several efforts have looked at diagnosing \emph{aortic stenosis} as the target outcome of a machine learning classifier, though none we are aware of use echocardiograms. 
\citet{yangClassificationAorticStenosis2020} used wearable sensors to build binary AS classifiers using data from 34 total subjects, reporting 96\% accuracy with random forests.
\citet{kwonDeepLearningBased2020} used \emph{electro}cardiogram (ECG) signals to build binary AS classifiers from over 30,000 patients at 2 hospitals in Korea.
Their best 12-lead models achieve 0.861 AUROC on external validation.
\citet{hataClassificationAorticStenosis2020} used ECGs from 700 patients in Japan.
Their 12-lead models achieve 84.2\% precision and 72.7\% recall on a heldout set.
In clinical practice, echocardiograms (ultrasound images) are the primary imaging modality used to assess the aortic valve.
To the best of our knowledge, there does not seem to be previous work on detecting aortic stenosis (AS) from echocardiograms, likely because echocardiograms contain complex data (video clips and Doppler recordings) that are not routinely annotated as part of clinical care. Acquiring labels for these images is prohibitively time consuming and expensive.
Among early efforts to automate AS screening, our study stands out for its focus on ultrasound (the clinical standard for diagnosing AS) and for showing how \emph{semi-supervised learning} can overcome limited labeled data.

\subsection{Related Work: Semi-supervised learning for echocardiography}

Work on \emph{semi-supervised} learning for echocardiography is still in early stages.
Previously,~\citet{madaniDeepEchocardiographyDataefficient2018} pursued semi-supervised classification of echocardiograms using \emph{generative adversarial networks} (GANs)~\citep{goodfellowGenerativeAdversarialNets2014}, including a 15-way view classification task using 267 patients as well as a diagnostic task for left ventricular hypertrophy (LVH).
All 2269 images from 455 patients were \emph{manually preselected} for a particular relevant view.
Impressively, they report over 92\% accuracy at the LVH diagnosis task using GANs; their view classifiers were similarly competitive.

Our work builds upon~\citet{madaniDeepEchocardiographyDataefficient2018} in three key ways.
First, we pursue diagnosis \emph{without manually preselecting relevant views}. Our setting matches what is needed in a real deployment where view labels would not be available.
Second, we develop methods to aggregate image-level predictions to produce \emph{patient-level} decisions, showing we can do much better than simple averaging.
Finally, we offer more streamlined and competitive methodology.~\citet{madaniDeepEchocardiographyDataefficient2018} suggest \emph{different} methods for semi-supervised and fully-supervised settings, using GANs for the former and convolutional neural nets (CNNs) for the later.
Instead, modern SSL methods can coherently pursue the same task no matter how much labeled data we have.
Our approach builds on well-established CNNs and extensions to wide residual architectures~\citep{zagoruykoWideResidualNetworks2017}.
We do not need the complexities of GANs or their well-known training difficulties~\citep{metzUnrolledGenerativeAdversarial2017,aroraGANsActuallyLearn2017}. 
Recent evaluations~\citep[Table 4]{miyatoVirtualAdversarialTraining2019} suggest the methods we build upon reduce error rates by 50\% over GANs on SSL benchmarks like SVHN.


%% file: fig_view_samples.tex
\newcommand{\BM}{0.3}
\setlength{\tabcolsep}{0.1cm}
\begin{figure}
\begin{tabular}{c c c c}
    PLAX  & PSAX & OTHER 
    \\
    \includegraphics[width=\BM\textwidth]{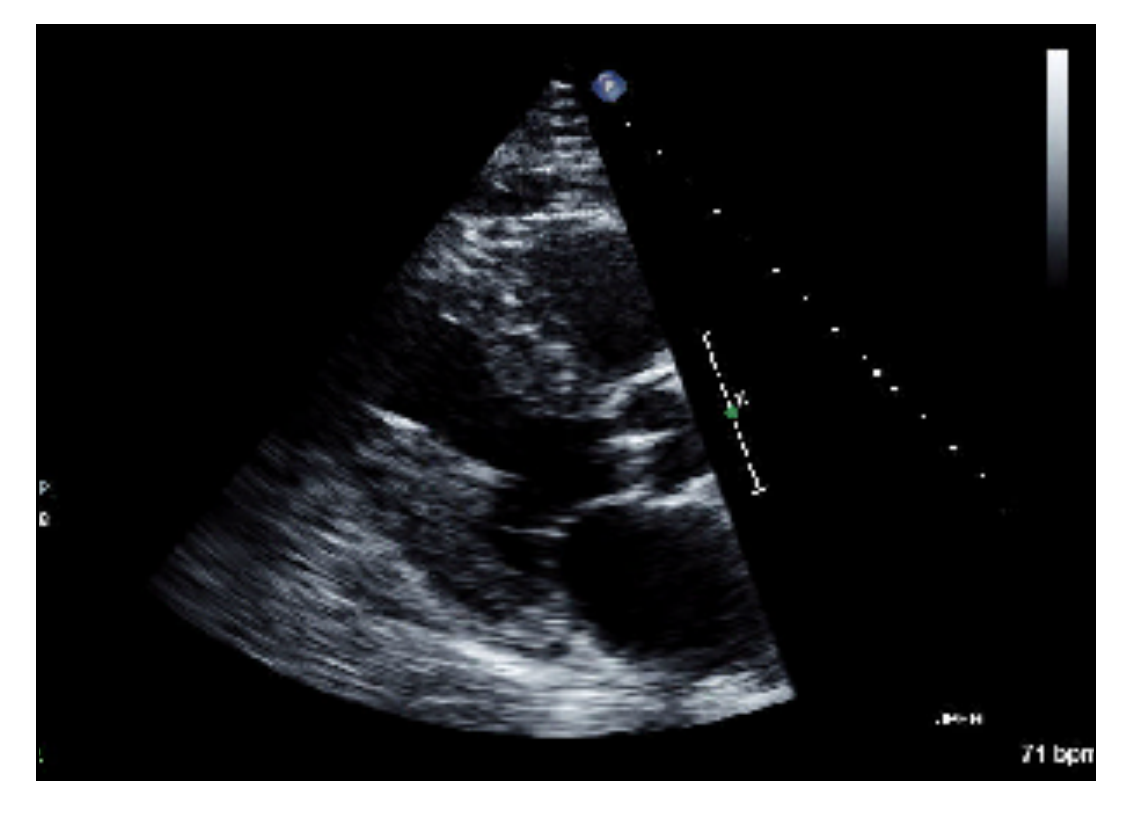}
    &
    \includegraphics[width=\BM\textwidth]{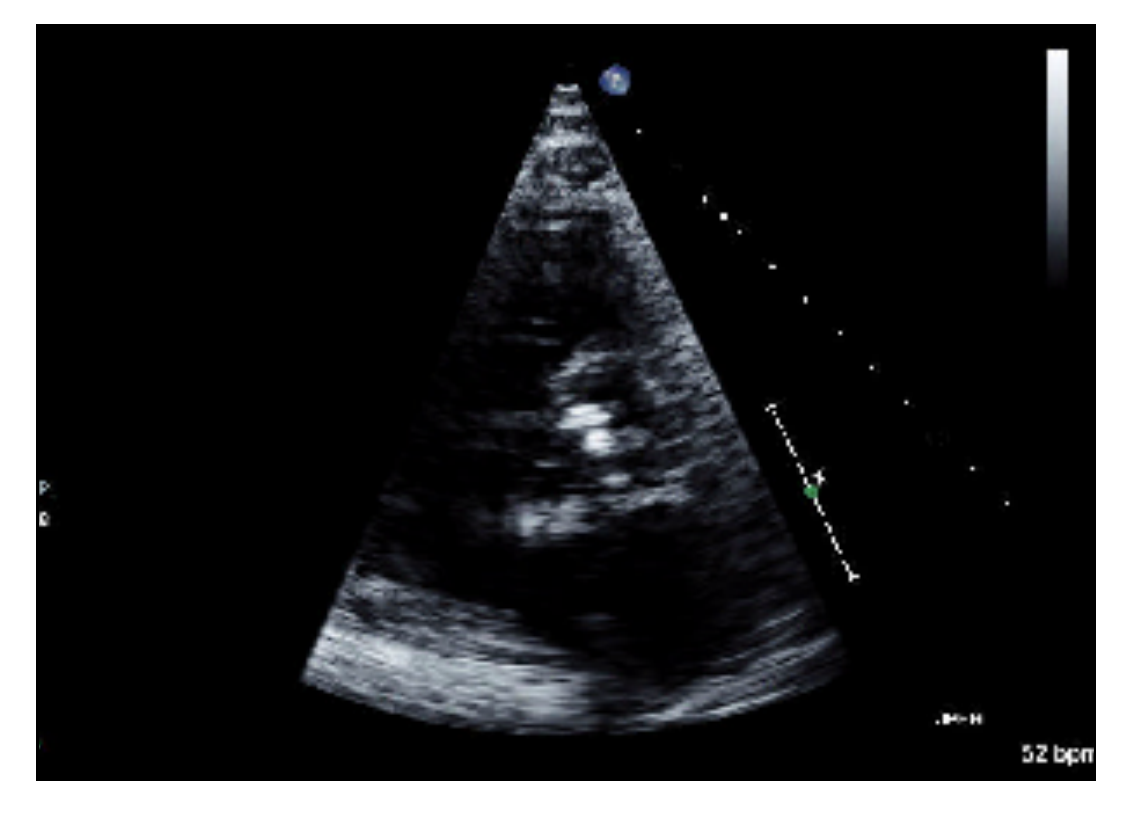}
    &
    \includegraphics[width=\BM\textwidth]{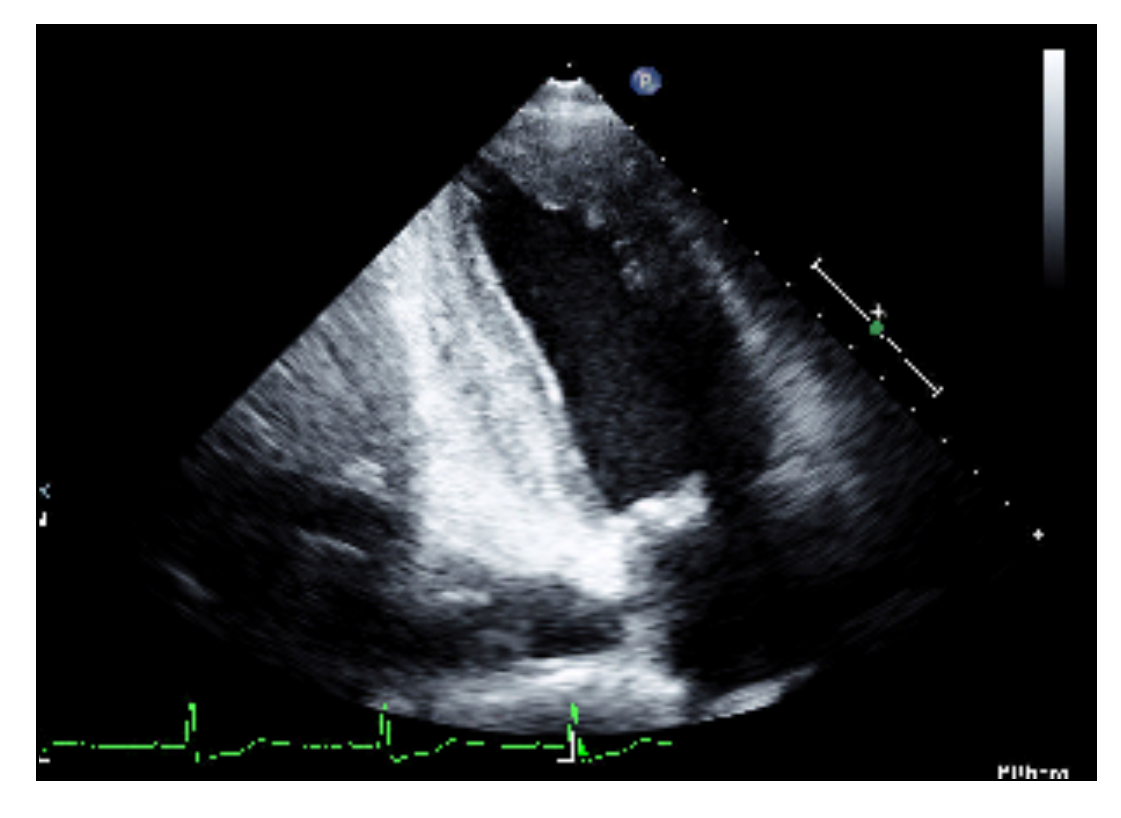}
    &
       
    \\
    
    \includegraphics[width=\BM\textwidth]{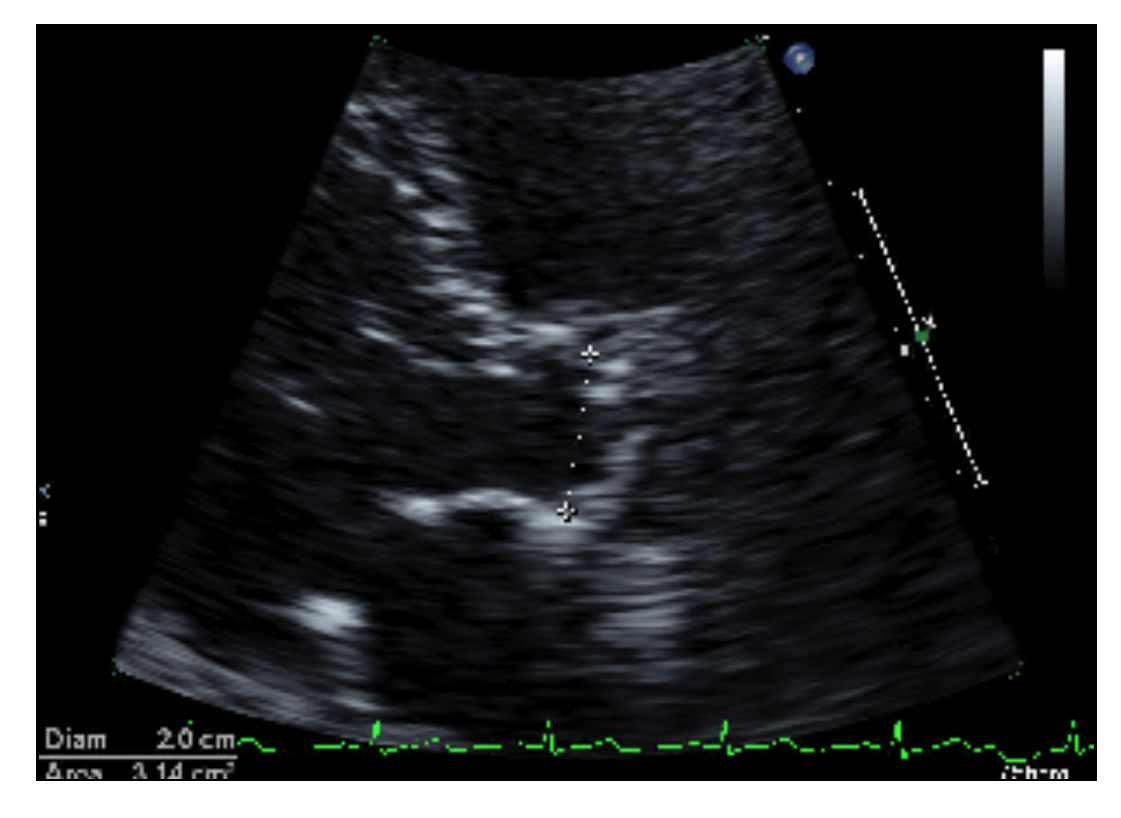}
    &
    \includegraphics[width=\BM\textwidth]{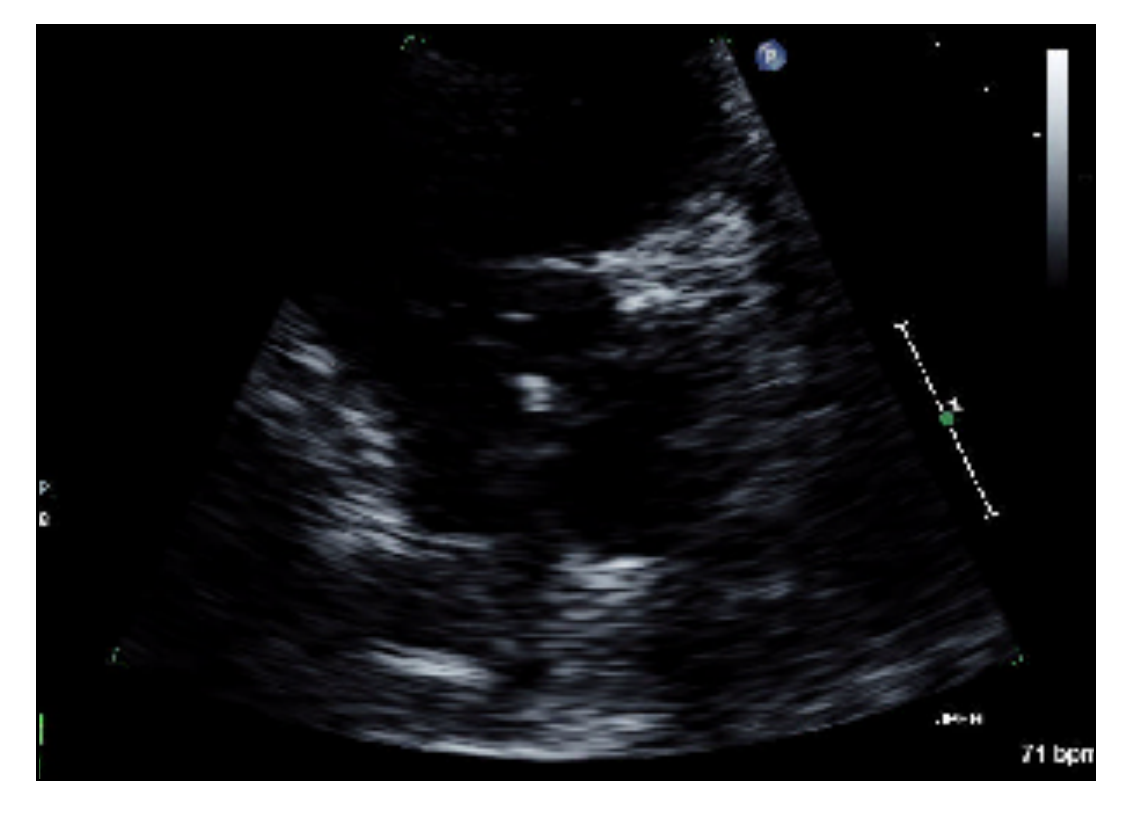}
    &
    \includegraphics[width=\BM\textwidth]{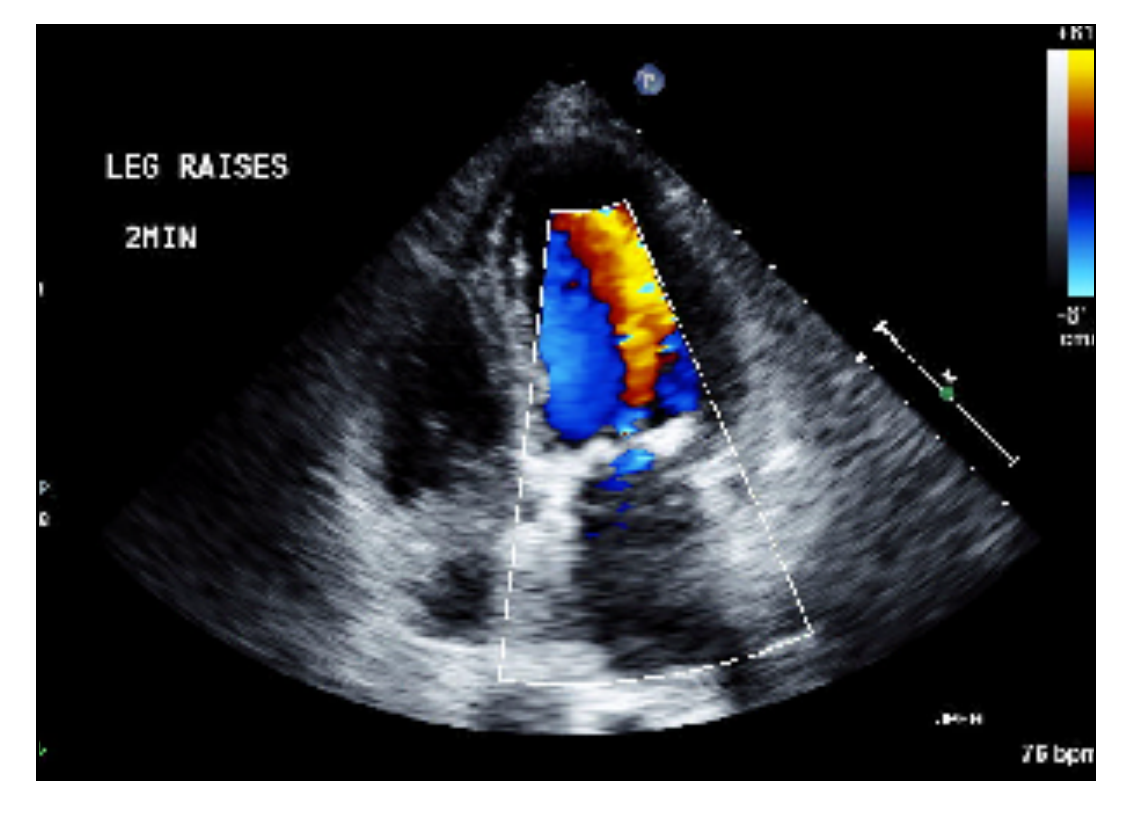}
    &
   
     \\
    
    \end{tabular}	
    \caption{Examples of images with their view label in our dataset. \emph{From left to right:} Two examples of parasternal long axis (PLAX) view, two examples of parasternal short axis (PSAX) view, and two examples of other kinds of view in our ``Other'' class. (Note: Images shown here are the original high-resolution images. When training the model, we further downsized the images to 64 x 64, see sec \ref{sec:image_preprocessing}) }
    \label{fig:VIEW_SAMPLES_MAIN}
\end{figure}

%% file: dataset.tex
\subsection{Dataset extraction and approval}

The dataset used in this paper contains a total of 2905 echocardiogram studies\footnote{Our released TMED dataset contains 2773 studies, each guaranteed to be from a distinct patient. The experiments here use a slightly larger dataset including multiple studies from the same patient. We have ensured each patient in the released data has exactly one study for simplicity.}. 
We use transthoracic echocardiogram (TTE) imagery acquired in the course of routine care consistent with American Society of Echocardiography (ASE)   guidelines~\citep{mitchellGuidelinesPerformingComprehensive2019}.
Each patient study contains multiple cineloop video clips of the heart depicting various anatomical views.
To collect this imagery, a sonographer manipulates a handheld transducer over the patient's chest, manually choosing different acquisition angles in order to fully assess the heart’s complex anatomy.
For this study we focus on still images extracted from all available video clips.
While pulsed-wave Doppler (PW), continuous-wave Doppler (CW), and m-mode recordings are also available, we leave these for future work.

The echocardiograms originate from the last 5 years of records at Tufts Medical Center, a high-volume tertiary care center in Boston, MA.
Echocardiograms are generally performed to assess for structural heart disease. These studies are done for a variety of reasons, from evaluating symptoms (e.g. chest pain, shortness of breath), to caring for a patient experiencing a cardiac event (e.g. myocardial infarction or acute heart failure), to providing follow up care for a known condition (e.g. aortic stenosis or cardiomyopathy). Studies were sampled from archived image folders that were organized by month of acquisition.

The use of these \emph{deidentified} images for research has been approved by our Institutional Review Board (Tufts IRB \#MODCR-03-12678).
All images were acquired from The Cardiovascular Imaging \& Hemodynamic Laboratory, part of the Tufts CardioVascular Imaging Center. 
This lab is Intersocietal Accreditation Commission (IAC) accredited and performs roughly 10,000 ultrasound examinations per year using devices from major vendors (Philips®, Toshiba®, Siemens®).
By using standardized image formats, the released data and subsequent ML methods are intended to be vendor-independent. 

\subsection{Image preprocessing.}
\label{sec:image_preprocessing}
Each study's raw imaging data contains multiple cineloop videos.
Typically there are around 100 - 200 videos per study.
From each cineloop file, we take one image to analyze.
Clinical collaborators suggested that any single frame could be used; in practice we took the first frame of each video.
The resulting data contains both color images and gray scale images with various resolutions. To prepare for neural network training, we convert each image to gray-scale, pad along its shorter axis to achieve a square aspect ratio, and then resize the image to 64x64 pixels. We filtered out Doppler recordings completely using aspect ratio since Doppler recordings have distinct aspect ratios. More data processing details are available in App.~\ref{sec:removing_doppler}.  

\subsection{Annotating each study with view and diagnostic labels}

As part of routine clinical care, there are no annotations applied to individual cineloops or still images for either view or diagnosis when imaging is collected.
Instead, images for a given study are reviewed in aggregate by an echocardiography-board-certified cardiologist to create a summary report that is merged into the electronic medical record.
This report contains diagnostic labels, including the presence or absence of AS and also the grade of AS, reliably collected for most patients.
All complete study reports produced in routine care have an assigned grade of aortic stenosis (range: no AS to severe AS).

However, while the diagnostic AS grade is available for all studies for which an expert reader has prepared a summary report, as implemented in our institution's current system it requires substantial manual effort to extract this label from the report into a form amenable to machine learning.
Furthermore, view labels are not available for any of the imagery we use.
Below, we detail how we obtain suitable annotations for a subset of patient-studies, which we call the \emph{labeled} set.
For the remaining studies in the \emph{unlabeled} set, we have only images: no view or diagnostic labels are easily available.

\paragraph{Diagnostic labels of AS disease severity.}
For this investigation, we were able to extract the AS grade from the relevant summary report in the EMR for a subset of all studies. We refer to this subset as the labeled set.
Each labeled patient in our study thus has an ordinal class label indicating one of 3 possible levels of severity: ``no AS'', ``mild/moderate AS'' and ``severe AS''. 
These patient-level diagnostic labels were assigned in standard fashion during routine care, integrating information across all available views for a given patient by a cardiologist with specialty training in echocardiography. 

We chose a 3-level granularity\footnote{
The raw data contains 5-level labels (our ``mild/moderate AS'' category can be further distinguished into ``mild'', ``mild to moderate'' and ``moderate''). We leave these finer-grained levels to future work.
} for AS severity classification -- no AS, mild/moderate AS, and severe AS --
as a good balance between simplicity and clinical utility.
In Supplement Fig. \ref{fig:PatientSevereAS} and Fig. \ref{fig:PatientNoAS}, we show example images from a patient with severe AS and a patient without AS; distinguishing these two categories is difficult to an untrained eye.

\paragraph{View labels.}
Because view labels were not available for any imagery, we undertook a significant post-hoc annotation effort.
We used a novel view labeling tool that displays a grid of multiple study images and facilitates rapid expert annotation. 
A board certified cardiologist provided all view label annotations.
Our expert annotator provided a view label for each image in our labeled set, selecting one of 3 possible labels: parasternal long axis (PLAX), parasternal short axis (PSAX) and a final category (Other) indicating all other possible view types that are not PLAX or PSAX.
We chose to focus on PLAX and PSAX view labels because the aortic valve's structure and function is visible. PLAX and PSAX views are used in the routine clinical assessment of aortic valve disease.
Fig.~\ref{fig:VIEW_SAMPLES_MAIN} shows examples of each view type; more samples can be found in Supplement Fig.~\ref{fig:VIEW_SAMPLES_APPENDIX}.

\paragraph{Summary of available labeled and unlabeled data.}
Out of all 2905 studies,
260 studies were assigned both view and diagnosis labels; 174 additional studies were assigned diagnosis but not view labels (while still difficult to automate in our current system, extracting a diagnosis severity label is easier than assigning a view label). The remaining 2471 studies are \emph{truly unlabeled}, with neither diagnosis nor view annotations available.

This data is further processed into two versions for standardized evaluation of SSL classification. The full-size version -- \datasetName-156-52 -- is described in Sec.~\ref{sec:data_full} and smaller version -- \datasetName-18-18 -- in Sec.~\ref{sec:data_small}.


%


\subsection{Full dataset with authentic unlabeled examples: \datasetName-156-52.}
\label{sec:data_full}

The full-size dataset used in this investigation consists of a \emph{labeled set} of all 260 fully-labeled patient studies (both view and diagnosis labels), as well as a much larger \emph{unlabeled set}. We review the design of each labeled and unlabeled set below.
All methods using the labeled set have access to diagnosis labels for each patient and view labels for each image in that set. No such labels are available in the unlabeled set.

\paragraph{Labeled train/valid/test sets.}
To evaluate the performance of classifiers on heldout data, we divide the labeled set of 260 patients using a 3:1:1 ratio into a labeled train set of 156 patients and evaluation sets (validation and test) of 52 patients each.
We call our full-size dataset the \datasetName-156-52 dataset, so that the true number of patients used for training and evaluation is apparent.
We repeat this partitioning 4 separate times, resulting in 4 independently-chosen train/valid/test sets, with summary counts in Table \ref{tab:patient_counts_large}. 

\paragraph{Unlabeled set.}
To build the full-size unlabeled set used to train semi-supervised methods, we combine the 2471 truly unlabeled patient-studies together with the 174 patient-studies that only have diagnosis labels (we discard any labels and treat them as unlabeled).
This results in a combined full-size unlabeled set with data from 2645 total patients. We have $\sim18x$ more unlabeled images than we have labeled training images.

\paragraph{Bonus heldout set for diagnosis.}
Because our available fully-labeled data is limited, to further assess diagnosis classifiers, we use the 174 studies with diagnosis labels as a \emph{bonus heldout set}. This use lets us evaluate if the rankings of methods on the original labeled test sets (52 patients) are repeatable in the larger 174 patient corpus.

\subsection{Smaller dataset: \datasetName-18-18.}
\label{sec:data_small}

Our full-size dataset described above contains labeled data from hundreds of patients. To simulate the practical scenario where we have access to only a few dozen labeled patient studies (e.g. in early prototyping of a medical imaging ML pipeline), we also perform experiments comparing SSL methods on a \emph{smaller} version of our dataset, where both labeled and unlabeled sets are significantly smaller than the full-size data described above.
Because it is easier to train methods on smaller datasets, this smaller version also allows us to evaluate many more methods on a fixed computational budget than our full-size dataset.

\paragraph{Labeled train/valid/test sets for smaller version.}
We select 54 patients to comprise the smaller labeled set, from the entire full labeled dataset of 260 patients.
Within the selected labeled set, we do a 1:1:1 train/validation/test split, favoring larger heldout size ratios here than in the full-size dataset to be sure we can assess real differences between models.
Thus, the \emph{labeled training set} contains data from 18 patients and each \emph{labeled heldout set} (validation and test) contains data from 18 patients (each patient's data is exclusively used for either training, validation, or test).
We call our smaller dataset the \datasetName-18-18 dataset (again to signify that 18 patient studies are available for training, and 18 for evaluation).
We repeat this partitioning 3 times, resulting in 3 independently-chosen train/valid/test partitions that balance the frequencies of each view and diagnostic label.
Summaries of each label's frequency are shown in Table~\ref{tab:image_counts_small}.

\paragraph{Unlabeled set for smaller version.}
To build the unlabeled set for \datasetName-18-18, we combine the remaining 206 patients from the full-size labeled set with the 174 patients that only have diagnosis labels.
Together, these 380 patients form the unlabeled set; even though we technically have labels for these studies, they are not used at all in training or evaluation.
In \datasetName-18-18, our unlabeled set has $\sim21x$ more images than the labeled train set.


\begin{table}[!t]
    \centering
    {\small 
    \begin{tabular}{l|r|rrr|r|rrr}
		& \multicolumn{4}{c}{Num. Patients}
		& \multicolumn{4}{c}{Num. Images}
	\\
    \textit{By Diagnosis}
    	& Total
    	& No AS  & Mild/Mod. & Severe    	
    	& Total
    	& No AS  & Mild/Mod. & Severe \\
    \hline
    Train (labeled) 
		& 18 & 6 & 6 & 6
    	& 1805, 1935
    	& 542, 635 & 634, 700 & 583, 647
	\\
    Valid. (labeled) 
    	& 18 & 6 & 6 & 6
    	& 1833, 2018
    	& 520, 663 & 586, 723 & 652, 704 
    \\
    Test (labeled) 
		& 18 & 6 & 6 & 6
    	& 1834, 1957
    	& 561 609 & 577, 668 & 691, 728
	\\ \hline
    \textit{By View}
		& & & &
    	& Total
    	& PLAX  & PSAX & Other
    \\ \hline
    Train (labeled) 
		& & & &
    	& 1805, 1935
    	& 208, 222 & 58, 72 & 1520, 1650
    \\
    Valid. (labeled) 
		& & & &
    	& 1833, 2018
    	& 212, 221 & 70, 81 & 1542, 1716
	\\
    Test (labeled) 
		& & & &
    	& 1834, 1957
    	& 215, 233 & 66, 77 & 1549, 1674
    \\
	\hline
    Unlabeled Train 
		& 380 & &  &
    	& 41183, 41428
    \end{tabular}
    }
    \caption{\textbf{Smaller \datasetName-18-18 dataset} summary, showing patient (left) and image (right) counts by diagnosis (top) and view (bottom).
    We use 3 different train/valid./test splits of the labeled set, all with same patient counts by diagnosis. To count images, we report (min., max.) values across splits.
    }
    \label{tab:patient_counts_small}
    \label{tab:image_counts_small}
\bigskip
    {\small 
    \begin{tabular}{l|r|rrr|r|rrr}
		& \multicolumn{4}{c|}{Num. Patients}
		& \multicolumn{4}{c}{Num. Images}
	\\
    \textit{Diagnosis}
    	& Total
    	& No AS  & Mild/Mod. & Severe    	
    	& Total
    	& No AS  & Mild/Mod. & Severe \\
    \hline
    Train
        & 156 & 49 & 47 & 60
    	& 16463, 16852
    	& 4681, 4801 & 5072, 5102 & 6589, 7076
    \\
    Valid.
	    & 52 & 16 & 16 & 20
    	& 5470, 5617
    	& 1458, 1544 & 1769, 1859 & 2153, 2330
    \\
    Test
	    & 52 & 16 & 16 & 20
    	& 5377, 5855
    	& 1542, 1636 & 1781, 1901 & 1931, 2443
	\\ \hline
    \textit{View}
		& & & &
    	& Total
    	& PLAX  & PSAX & Other
    \\ \hline
    Train
		& & & &
    	& 16463, 16852
    	& 1883, 1918 & 595, 672 & 13908, 14297
    \\
    Valid.
		&  & &  &
    	& 5470, 5617
    	& 639, 652 & 184, 221 & 4644, 4749
	\\
    Test
		& & & &
    	& 5377, 5855
    	& 624, 646 & 172, 212 & 4559, 5027
    \\
    \hline
    Unlabeled & 2645 &  &  &  & 303498
    \end{tabular}
    }
    \caption{\textbf{Full-size \datasetName-152-52 dataset} summary, showing patient (left) and image (right) counts by diagnosis (top) and view (bottom).
    We use 4 different train/valid/test splits of the labeled set, all with same patient counts by diagnosis. To count images, we report (min., max.) across splits.
    }    
    \label{tab:image_counts_large}
	\label{tab:patient_counts_large}
\end{table}



%% file: methods.tex
\subsection{Methods for SSL Image Classification}
\label{sec:Methods-SSL}

Inspired by \citet{oliverRealisticEvaluationDeep2018}, we wish to carefully evaluate semi-supervised learning methods for image classification, focusing on a simple SSL baseline (Pseudo-Label) as well as two recent high-performing methods: virtual adversarial training (VAT, \citep{miyatoVirtualAdversarialTraining2019}) and MixMatch~\citep{berthelotMixmatchHolisticApproach2019}. Below we review the key ideas behind how each method learns from labeled and unlabeled data.
All descriptions below use the neural network notation defined in Sec.~\ref{sec:ssl_background}.
Hyperparameters and other settings are found in App.~\ref{sec:arch_and_hyperparameters}.

\paragraph{Pseudo-Label.} The pseudo-label method~\citep{leePseudolabelSimpleEfficient2013} is a natural way to use unlabeled data to help train a neural network.
At each minibatch of unlabeled data during stochastic gradient descent, we use the existing classifier to make predictions,
 obtaining the \emph{pseudo-label} $\hat{y}(x) \in \mathcal{C}$ indicating the most likely predicted class for image $x$.
If the predicted probability of the most likely class is above a user-specified \emph{confidence threshold} $\tau$, we include this example in the loss, treating the pseudo-label as the true label.
Thus, for our tasks the Pseudo-Label unlabeled loss $\mathcal{L}^U(x, \theta)$ is either weighted cross-entropy $\ell$ or zero (if the image is excluded):
\begin{align}
    \mathcal{L}^U(x, \theta) &= 
		1[ S(f_{\theta}(x))_{\hat{y}(x)} > \tau ]
    	\cdot \ell( \text{one-hot}(\hat{y}(x)), S(f_{\theta}(x))
, \quad  \hat{y}(x) = \argmax_{c \in \mathcal{C}} f_{\theta}(x)_c
\end{align}
where $1[ \cdot ]$ is an indicator function that returns either one (if the expression is true) or zero.



\paragraph{Virtual Adversarial Training (VAT).}
Recently, \citet{miyatoVirtualAdversarialTraining2019} present \emph{virtual adversarial training} as a way to improve robustness for both supervised and semi-supervised classifiers.
The key idea is that for each image $x$ we can easily find a nearby \emph{perturbed} version of the image $x' = x + \Delta^*$, where $\Delta^*$ is the vector that leads to greatest change in predicted label distribution.
To achieve \emph{smooth} and \emph{consistent} predictions, we wish to penalize cases where the predictions for $x$ differ from those for the nearby $x'$ (in KL divergence).
Every training image $x$ (both labeled and unlabeled) is assessed for this loss:
\begin{align*}
\mathcal{L}^U(x, \theta) &= 
    	\KL\left( S(f_{\theta}(x)), S(f_{\theta}(x + \Delta^*)) \right),
\quad 
\Delta^* = \argmax_{\Delta : || \Delta ||^2 \leq \epsilon }
	\KL( S(f_{\theta}(x)), S(f_{\theta}(x+\Delta)))
\end{align*}
The perturbation vector $\Delta^*$ is constrained to have magnitude below a given perturbation size $\epsilon > 0$. Its value can be found efficiently using routines from~\citet{miyatoVirtualAdversarialTraining2019}.

\paragraph{MixMatch.} MixMatch~\citep{berthelotMixmatchHolisticApproach2019} learns from unlabeled data by combining two key ideas: data augmentation and a variation of pseudo-label's unlabeled loss function.

\textit{How MixMatch performs augmentation.}
MixMatch uses the unlabeled set as a key input to its data augmentation procedure. 
The core of this procedure is MixUp~\citep{zhangMixupEmpiricalRisk2017}, which \emph{linearly interpolates} between two given images.
During training, MixMatch visits each minibatch (which contains labeled and unlabeled data). 
Each source labeled image (and its label) is transformed via MixUp with a randomly selected other image-label pair in that minibatch.
If an unlabeled example is selected for pairing, we create a pseudo-label $q(x)$ from the probabilistic vector output of the classifier: $q(x) = S( f_{\theta}(x))$. 
The resulting transformed labeled minibatch is fed into the labeled loss.
Thus, unlike other SSL methods described above, unlabeled data can inform training via augmentation alone, even if the unlabeled loss $\mathcal{L}^U$ is omitted.


\textit{How MixMatch calculates unlabeled loss.}
MixMatch also transforms each unlabeled image $x$ in a  minibatch via MixUp to obtain $x'$, mixing with either labeled or unlabeled images.
These transformed examples are fed into a pseudo-label inspired unlabeled loss:
\begin{align}
\mathcal{L}^U(x, x', \theta) &= \frac{1}{C} \sum_{c=1}^{C} \left( q(x)_c - S(f_{\theta}(x'))_c \right)^2.
\end{align}
Unlike the original pseudo-label method, within MixMatch the pseudo-label $q(x)$ is a probability vector (rather than a one-hot vector).
Overall, this loss has \emph{no} threshold-based exclusion, so all unlabeled examples can contribute to the loss, and uses mean squared error (rather than alternatives like KL) as recommended for robustness by~\citet{berthelotMixmatchHolisticApproach2019}.

\paragraph{Ablation: Augmentation-Only MixMatch.}
Given MixMatch's complexity, a natural question arises: does MixMatch benefit from unlabeled data because it informs the labeled loss via augmentation or because of the unlabeled loss directly?
While the original work did many other ablations~\citep{berthelotMixmatchHolisticApproach2019}, this question was not directly answered.
Our experiments thus assess a variant called \textbf{Augment-Only MixMatch}, which omits the unlabeled loss but still uses unlabeled data to inform the labeled loss via augmentation.


\subsection{Standardized architecture, training, and hyperparameter selection.}
\label{sec:standard-SSL-framework}

Our main methodological interest is assessing \emph{how much} the SSL paradigm improves task performance on our datasets by incorporating unlabeled data, as well as \emph{which techniques} might work the best, as to our knowledge methods like VAT and MixMatch have not yet been applied to echocardiograms.
To put all methods on a fair footing, we use a common architecture (a WideResNet) and develop a standardized protocol for training parameters (ADAM with modest L2 regularization) used by all methods.
All hyperparameters are selected via a grid search to maximize balanced accuracy on the validation set.
Details about all architectures, hyperparameters, and training procedures are available in App.~\ref{sec:arch_and_hyperparameters}.
Our open-source code provides everything needed for reproducing our implementation\footnote{\codeURL}.

Below, we highlight two implementation choices that give consistent gains to all methods.

\paragraph{Unlabeled loss weight hyperparameter.}
For every SSL method we study, the unlabeled loss weight hyperparameter $\lambda > 0$ matters.
We follow the recommendations of \citet{leePseudolabelSimpleEfficient2013} and use a \emph{deterministic annealing} schedule to slowly increase $\lambda$ over epochs from an initial value of 0.0 to its maximum value linearly over the course of many training iterations. 
We select the maximum value of $\lambda$ for all methods by monitoring the model performance on validation set, and select the $\lambda$ that gives best validation performance.
A well-chosen unlabeled loss schedule is important to achieve good performance for MixMatch and other methods.
Supplementary Table~\ref{tab:MixMatch hyperparameters ablation study} suggests that tuning the schedule can improve balanced accuracy by over 1\%.

\paragraph{Ensembling models over one training run to improve generalization.}
A recent study by \citet{huangSnapshotEnsemblesTrain2017} suggests that rather than using the final checkpoint of an SGD training run, or even the best single checkpoint as ranked by validation loss, an ensemble of the checkpoints along the training trajectory can achieve better generalization.
We apply this ensemble method to \emph{every method} (SSL and baseline labeled-set only methods). We further incorporate the \emph{weighted average} idea from~\citet{caruanaEnsembleSelectionLibraries2004} , allowing better-performing checkpoints larger influence.
The final performance of each method is determined via an \emph{ensemble} of the last 25 checkpoints (one per epoch).
Supplement Table~\ref{tab:best_single_checkpoint_VS_ensemble_FS_echo260} shows that this ensembling improves performance by a modest but noticeable amount.


\subsection{Methods for aggregating many images into patient-level predictions}
\label{sec:methods-patient-level}

A key aspect of our study is producing useful diagnosis predictions for a specific patient (indexed by $n$), based on multiple echocardiogram images collected for that patient (indexed by $i \in \{1, 2, \ldots I_n\}$). Each patient may have a different number of images $I_n$, with typical $I_n$ values in the 100s.
Given all images, we wish to predict the \emph{patient's} diagnosis label $y_n$ (one of no AS, mild/moderate AS, or severe AS).
We will use the image-specific view classifier network $f_{\theta_V}$ and diagnosis classifier network $g_{\theta_D}$ introduced in Sec.~\ref{sec:ssl_background}.

Below, we present several strategies for aggregating probabilistic predictions from several images to produce a \emph{patient-level} prediction.
Previous studies have often manually prescreened a subset of images whose view types are known to be relevant to the prediction tasks. Only these prescreened images are used to make patient-level predictions.
Instead, we consider the task faced in a real deployment where no manual prescreening is available, and we must compute the probability of the patient's diagnosis from all $I_n$ images: $p( y_n | x_{n, 1:I_n} )$.


\paragraph{Simple average.} One aggregation strategy is to simply average over the diagnosis predictions for each of the $I_n$ images available for patient $n$, treating each image equally:
\begin{align}
p(y_n = c | x_{n,1:I_n} ) = 
	 \frac{1}{I_n}
	 \sum_{i=1}^{I_n} S(g_{\theta_D}(x_{ni}))_c.
\end{align}
While simple and used in previous work~\citep{ghorbaniDeepLearningInterpretation2020}, this method will be error-prone if many images do not depict anatomical features relevant to AS diagnosis.

\paragraph{Prioritize diagnoses from relevant views.}
To diagnose AS, the PLAX and PSAX views show the anatomical structures that are \emph{relevant}, while our catch-all ``Other'' view  contains many diverse view types that are mostly (but not completely) irrelevant.
Our view classifier network (with weights $\theta_V$) can predict which images depict a \emph{relevant} view (PLAX or PSAX).
Thus, we suggest an aggregation procedure for diagnosis predictions that uses a \emph{weighted} average over images.
Each image's weight $w(x_{ni}) \in (0,1)$ is the view classifier's probabilistic confidence that the image shows a \emph{clinically-relevant} view for our task:
\begin{align}
p(y_n = c | x_{n,1:I_n} ) \propto
\sum_{i=1}^{I_n} w(x_{ni}) S(g_{\theta_D}(x_{ni}))_c, \quad w(x_{ni}) = p( v_{ni} \in \mathcal{R} | x_{ni}, \theta_V).
\end{align}
Here, the set of \emph{relevant} view types $\mathcal{R}$ contains PLAX and PSAX but not ``Other.''

We explored an alternative strategy of prioritization which \emph{thresholds} to identify a subset of relevant images (all treated equally) rather than probabilistic weighting in which all images contribute proportional to their weight. We found this strategy's performance is slightly inferior to our weighting strategy. Details can be found in Appendix \ref{tab:Suggested_Aggregation_Ablation}.

\paragraph{Learned image-to-patient prediction function.} 
We can further imagine \emph{training a model} that can produce the predicted probabilities needed for a patient, given relevant features from its component images.
We explored a few possible approaches based on manually engineered features and logistic regression classifiers, but did not find these delivered benefits worth the extra implementation effort. We leave this idea as a possible future direction.



\subsection{Transfer from view to diagnosis by pretraining}
\label{sec:transfer_from_view_to_diagnosis}

A unique property of our problem and dataset is that we have two types of labels for the same input: view labels and diagnosis labels.
We further have clinical knowledge that the tasks are closely related (successful diagnosis requires the ability to identify relevant views).
We leverage this relation to improve training of our diagnosis classifiers.
Specifically, we \emph{pretrain} a single-image view classification network and then use this network's weights as a warm-start for our diagnosis classifier.
Note this is different from common transfer learning practice where a network is pretrained on some other dataset.
Our method does not require an additional dataset, merely other labels on the same dataset.

\subsection{Multi-task learning using both diagnosis and view labels}
Another way to improve diagnosis using view labels is via multitask learning~\citep{ruder2017overview,zhang2021survey}, thought it is sometimes challenging in practice to determine whether auxiliary tasks will be helpful or harmful to the main task \citep{zhang2021survey, ruder2017overview}.
We investigate whether a simple multi-task approach that trains the same network to jointly recognize view and diagnosis could be effective. 
Our multi-task labeled loss is:
\begin{align}
\mathcal{L}^L(\theta) 
&= \ell( y, S(g_{\theta_D}(x))) + \gamma \ell( v, S(f_{\theta_V}(x))), 
\label{eq:multitask}
\end{align}
where $y$ represents the image's one-hot diagnosis label, $v$ is the one-hot view label, and $\ell$ is a weighted cross-entropy loss.
Hyperparameter $\gamma$ controls the strength of the view loss, as we ultimately care most about diagnosing the AS severity.
We focus on \emph{labeled-set-only} evaluation of this strategy. Future work could explore semi-supervised multi-task learning.



%% file: results.tex
To evaluate whether state-of-the-art SSL can improve real-world echocardiogram classification, we now compare several SSL methods against a baseline network of the same architecture that learns from only the labeled training set.
We compare Pseudo-label \citep{leePseudolabelSimpleEfficient2013}, virtual adversarial training~\citep{miyatoVirtualAdversarialTraining2019}, and MixMatch \citep{berthelotMixmatchHolisticApproach2019}. 

We first investigate image-level view classification in Sec.~\ref{sec:results-view}, then image-level diagnosis classification in ~Sec.~\ref{sec:results-diagnosis-from-images}.
Finally, in Sec.~\ref{sec:results-diagnosis-from-patient} we investigate how well our methods perform at patient-level diagnosis, using the image aggregation strategies from Sec.~\ref{sec:methods-patient-level}.

\paragraph{Performance metric.} For all view and diagnosis tasks, we use \emph{balanced accuracy} as our primary performance metric.
Given a dataset of $N$ true labels $y_{1:N}$ and $N$ predicted labels $\hat{y}_{1:N}$, with each label $y_n \in \{1, \ldots C\}$, we compute balanced accuracy as
\begin{align}
\text{balanced-accuracy}(y_{1:N}, \hat{y}_{1:N}) &= \frac{1}{C} \sum_{c=1}^{C} \frac{\text{TP}_{c}(y_{1:N}, \hat{y}_{1:N})}{N_{c}(y_{1:N})}.
\label{eq:balanced_accuracy}
\end{align}
Let $\text{TP}_c(\cdot)$ count \emph{true positives} for class $c$ (that is, the number of correctly classified examples whose true label is $c$), and $N_c(\cdot)$ gives the total number of examples with true label $c$.

We select balanced accuracy because standard accuracy does not adequately reflect performance on tasks with \emph{label imbalance}. In our view classification test set, the ``Other'' category is far more common, representing over 80\% of all images. Trivally guessing ``Other'' for every image would thus reach over 80\% accuracy, but only 33.3\% balanced accuracy.

\subsection{View classification}
\label{sec:results-view}

We first compare all selected methods on the small \datasetName-18-18 dataset in Table~\ref{tab:view classification small}.
All SSL methods provide gains over methods that only use the labeled set. The largest gains come from MixMatch, which improves the baseline by over 9\% in balanced accuracy.


Next, we study the best performing methods on the larger \datasetName-156-52 dataset in Table~\ref{tab:view classification large}. 
Again, we see visible gains from MixMatch over the labeled-set-only baseline, improving over 2.5\% in balanced accuracy. The relative gain of SSL is smaller here because the amount of labeled training data is larger (eventually, with enough labeled data the performance of all methods should saturate).
Since Pseudo-Label and VAT perform worse than MixMatch in our experiments on the smaller TMED-18-18 dataset, we did not assess these methods on the larger dataset to keep computation costs low.

We further observe that our simpler variant of MixMatch that only uses unlabeled data for augmentation, which we called Augment-Only MixMatch, captures most of the gains of MixMatch (around \textbf{74$\%$} for both datasets), suggesting that augmentation (rather than a well-designed unlabeled loss) is the primary reason for MixMatch's success.


\begin{table}[!h]
    \centering
    \begin{tabular}{r r l|ccc|c}
    \multicolumn{2}{c}{Number of \emph{Unlabeled}}
    	& 
    	& \multicolumn{3}{c}{Split-specific Test Set}
    \\
    Patients & Images & Method & 0  & 1 & 2 & average\\
    \hline
    0 & 0 & Basic WRN & 81.37 & 81.84 & 82.70 & 81.97\\
    380 & $\sim$41,000  & Pseudo-Label & 81.81 & 85.07 & 85.82 & 84.23\\
    380 &  $\sim$41,000  & VAT & 87.95 & 87.61 & 86.36 & 87.31\\
    380 &  $\sim$41,000  & Augment-Only MixMatch  & 87.77 & 92.43 & 86.06 & 88.75\\
    380 &  $\sim$41,000 & MixMatch & 92.11 & 93.07 & 88.15 & \textbf{91.11}
    \end{tabular}
    \caption{View classification on our \textbf{smaller \datasetName-18-18} dataset, showing test-set balanced accuracy across 3 separate train/test splits each with 18 labeled patients for training and 18 for test.
    }
    \label{tab:view classification small}
\bigskip

    \begin{tabular}{r r l|cccc|c}
    \multicolumn{2}{c}{Number of \emph{Unlabeled}}
    	& 
    	& \multicolumn{4}{c}{Split-specific Test Set}
    \\
    Patients & Images & Method & 0  & 1 & 2 & 3 & average\\
    \hline
	0 & 0 & Basic WRN & 92.37 & 93.24 & 93.72 & 93.87 & 93.30\\
    2645 & 303726 & Augment-Only MixMatch & 95.50 & 95.62 & 95.00 & 94.66 & 95.20\\
    2645 & 303726  & MixMatch & 96.22 & 95.79 & 95.77 & 95.74 & \textbf{95.88}
    \end{tabular}
    \caption{View classification on our \textbf{full-size \datasetName-156-52} dataset, showing test-set balanced accuracy across 4 train/test splits, each with 156 labeled patients for training and 52 for test.
        For simplicity, we focus on the best SSL method (MixMatch) identified in the smaller data comparisons.
    }
    \label{tab:view classification large}
\end{table}

\subsection{Diagnosis classification from a single image}
\label{sec:results-diagnosis-from-images}

To examine how SSL might improve AS diagnosis when given a single image, we first compare all candidate methods on the smaller dataset in Table~\ref{tab:diagnosis classification small}.
Like in the view task, we see that MixMatch beats all other SSL methods and the baseline by a substantial margin.
Further experiments on the full-size dataset in Table~\ref{tab:diagnosis classification large} again show MixMatch improves accuracy.

We also assess the added-value of our proposed pretrain-on-view transfer learning methods (Sec.~\ref{sec:transfer_from_view_to_diagnosis}).
Both Table~\ref{tab:diagnosis classification small} and Table~\ref{tab:diagnosis classification large} suggest that this pretraining strategy offers gains over simply initializing the weights of our single-image diagnosis classifier from scratch (across splits we average +3.4\% on the smaller dataset and +0.5\% on the larger dataset).
Notably these gains are \emph{consistent} across splits: we see a gain from pretraining visible in each of the 4 train/test splits.

In addition to pretraining, multitask learning also seems effective. We can see an average balanced accuracy gain of +4.4\% on the smaller dataset and +2.2 \% on the larger dataset. This demonstrates the added benefit of utilizing both view and diagnosis labels (not just diagnosis alone) to improve generalization performance.


\begin{table}[!h]
    \centering
    \begin{tabular}{r r l l|ccc|c}
    \multicolumn{2}{c}{Number of \emph{Unlabeled}} &
    	& &
    	\multicolumn{3}{c}{Split-specific Test Set}
    \\
    Patients & Images & Pretrain & Method & 0  & 1 & 2 & average\\
    \hline
    0 & 0 & &  Basic WRN & 42.25 & 51.26 & 36.73 & 43.41\\
    0 & 0 & &  Multitask WRN, & 43.46 & 54.30 & 45.80 & 47.85\\
    0 & 0 & view & Basic WRN & 41.76 & 52.21 & 40.40 & 44.79\\
    380 & $\sim$41,000 & & Pseudo-Label & 41.16 & 49.93 & 42.94 & 44.68\\
    380 & $\sim$41,000 & & VAT & 42.05 & 51.13 & 42.79 & 45.32\\
    380 & $\sim$41,000 & & MixMatch & 43.89 & 55.17 & 44.76 & 47.94\\
    380 & $\sim$41,000 & view & MixMatch & 50.01 & 56.46 & 47.73 &\textbf{ 51.43} 
    \end{tabular}
    \caption{AS severity diagnosis classification for \emph{individual images} on the \textbf{smaller \datasetName-18-18} dataset, showing balanced accuracy averaged over the test sets from multiple train/test splits of the labeled set.
     Each split's test set contains all images from 18 patients, and each split's labeled training set contains all images from 18 patients.
    The labeled-only Basic WRN transfers from corresponding Basic WRN view classifier, while Mixmatch transfers from corresponding MixMatch view classifier}
    \label{tab:diagnosis classification small}
\bigskip 
    \begin{tabular}{r r l l| rrrr | r}
    \multicolumn{2}{c}{Number of \emph{Unlabeled}}
        & & & \multicolumn{4}{c}{Split-specific Test Set}
    \\
    Patients & Images 
    & Pretrain & Method & 0 & 1 & 2 & 3 & average\\
    \hline
    0 & 0 & & Basic WRN & 62.95 & 61.03 & 56.58 & 63.84 & 61.13\\
    0 & 0 & & Multitask WRN & 66.16 & 62.41 & 58.70 & 66.98 & 63.31\\
    2645 & 303726 & & MixMatch & 65.57 & 62.69 & 60.87 & 66.29 & 63.86\\
    2645 &  303726& view & MixMatch & 67.39 & 62.79 & 61.02 & 67.36 & \textbf{64.64}\\ 
    \end{tabular}
    \caption{AS severity diagnosis classification for \emph{individual images} on the \textbf{full-size \datasetName-156-52} dataset, showing balanced accuracy across multiple splits. Each split's test set contains all images from 52 patients, and each split's labeled training set contains  all images from 152 patients.
    }
    \label{tab:diagnosis classification large}
\end{table}

\subsection{Diagnosis performance using all images related to a patient}
\label{sec:results-diagnosis-from-patient}

We finally consider how SSL methods might improve AS diagnosis when making predictions for a patient, aggregating information from many images with diverse view types, using the methods from Sec.~\ref{sec:methods-patient-level}.
The results on the full-size dataset in Table~\ref{tab:diagnosis classification large patient} suggest that SSL with MixMatch, when combined with our other key innovations (prioritizing relevant views, pretraining on view) offers real value, achieving 90.1\% balanced accuracy compared to the baseline's 81.57\%. Ablations in Table~\ref{tab:diagnosis classification large patient} help further disentangle how each piece (adding semi-supervised learning, adding prioritization, adding pretraining) help. The results suggest that prioritization of relevant views offers the largest and most consistent gains, followed by the semi-supervised learning, and then pretraining on view classification.

To further understand the source of these gains, we examine \emph{confusion matrices} in Fig \ref{fig:confusion_matrix} across 4 independent train/test splits of our full-size TMED-156-52 dataset.
This figure compares side-by-side our best classifier (pretrained MixMatch that prioritizes relevant views) and a labeled-set-only baseline using a simple average aggregation strategy.
We see that our proposed method consistently makes fewer mistakes across all splits: 3 fewer mis-diagnosed patients on split 1, 3 fewer on split 2, 4 fewer on split 3, and 6 fewer on split 4.  
For every severity level and split, the proposed method achieves equal or better recall than the baseline.

Our dataset is limited: even the full-size dataset has only 52 patients in the test set to evaluate results. 
Therefore, to better assess the significance of our claims (that SSL learning with MixMatch delivers improved performance, which is further boosted by smart prioritization of relevant views), we revisit the portion of our large dataset that had only diagnosis labels (and no view labels) for 174 patients. For this experiment, we call this the ``bonus heldout set''.
Results comparing all methods on this bonus heldout set are in Supplementary Table~\ref{tab:diagnosis classification patient unlabeled_heldout_174}.
We find our claims are consistent: among methods that use simple averaging, MixMatch improves over the Basic WRN baseline by over 1\% balanced accuracy, while when using prioritized views MixMatch improves by over 3.5\%.
We do note that these ``bonus set'' images were included in the \emph{unlabeled} training set. However, we emphasize that their labels were \emph{not used} during training and that they make up less than 6\% of the total unlabeled set.

\subsection{ROC analysis of No AS vs Some AS}
Finally, we assess our method's ability as a first-line screening tool by reporting a receiver operating curve in Supplementary Fig. \ref{fig: No AS vs Some AS}. for the simpler binary task of distinguishing between ``no AS'' and ``some AS'' (combining the ``mild/moderate AS'' and ``severe AS'' labels).
We compared a labeled-set-only model with simple averaging, a labeled-set-only model with prioritized voting, and our SSL-trained MixMatch method with prioritized voting.
While all compared methods achieve high area-under-the-ROC scores, we find that prioritized voting consistently shows gains, achieving a remarkable average AUC of \textbf{0.98} across the 4 splits.
This performance suggests that we may not be far from effective deployment of these models as a first-line screening tool, provided we can replicate this performance in future external validation.

\begin{table}[!t]
    \centering
    \begin{tabular}{l l l|rrrr|c}
	         &        & & \multicolumn{4}{c}{Split-specific Test Set} \\
    Pretrain & Method & Aggregation across images
    & 0  & 1 & 2 & 3 & average\\
    \hline
    & Basic WRN & Simple average & 87.92 & 81.67 & 72.92 & 83.75 & 81.57\\
    & Basic WRN & Prioritize relevant view & 90.83 & 80.00 & 83.33 & 92.50 & 86.66\\
    \hline
    & MixMatch & Simple average & 90.00 & 77.08 & 80.83 & 94.17 & 85.52\\
    & MixMatch & Prioritize relevant view & 88.75 & 83.75 & 85.00 & 94.17 & 87.92\\
    view & MixMatch & Simple average & 87.50 & 77.08 & 76.67 & 85.83 & 81.77\\
    view & MixMatch & Prioritize relevant view  & 93.75 & 87.50 & 82.92 & 96.25 & \textbf{90.11}\\
    \end{tabular}
    \caption{Patient-level AS severity diagnosis classification on the \textbf{full-size \datasetName-156-52 dataset}, showing balanced accuracy on the test set across multiple train/test splits. Each split's labeled training set contains 156 patients, and each test set contains 52 patients.
    MixMatch methods are trained with \emph{semi-supervised} learning on both labeled and unlabeled data. 
    }
    \label{tab:diagnosis classification large patient}
\end{table}

%% file: fig_confusion_mat.tex
\newcommand{\BW}{0.23}
\setlength{\tabcolsep}{0.1cm}
\begin{figure}
\begin{tabular}{r c c c c}
    & Split 1 & Split 2 & Split 3 & Split 4
    \\
    {\rotatebox{90}{~~~~~~Basic WRN}}
    & 
    \includegraphics[width=\BW\textwidth]{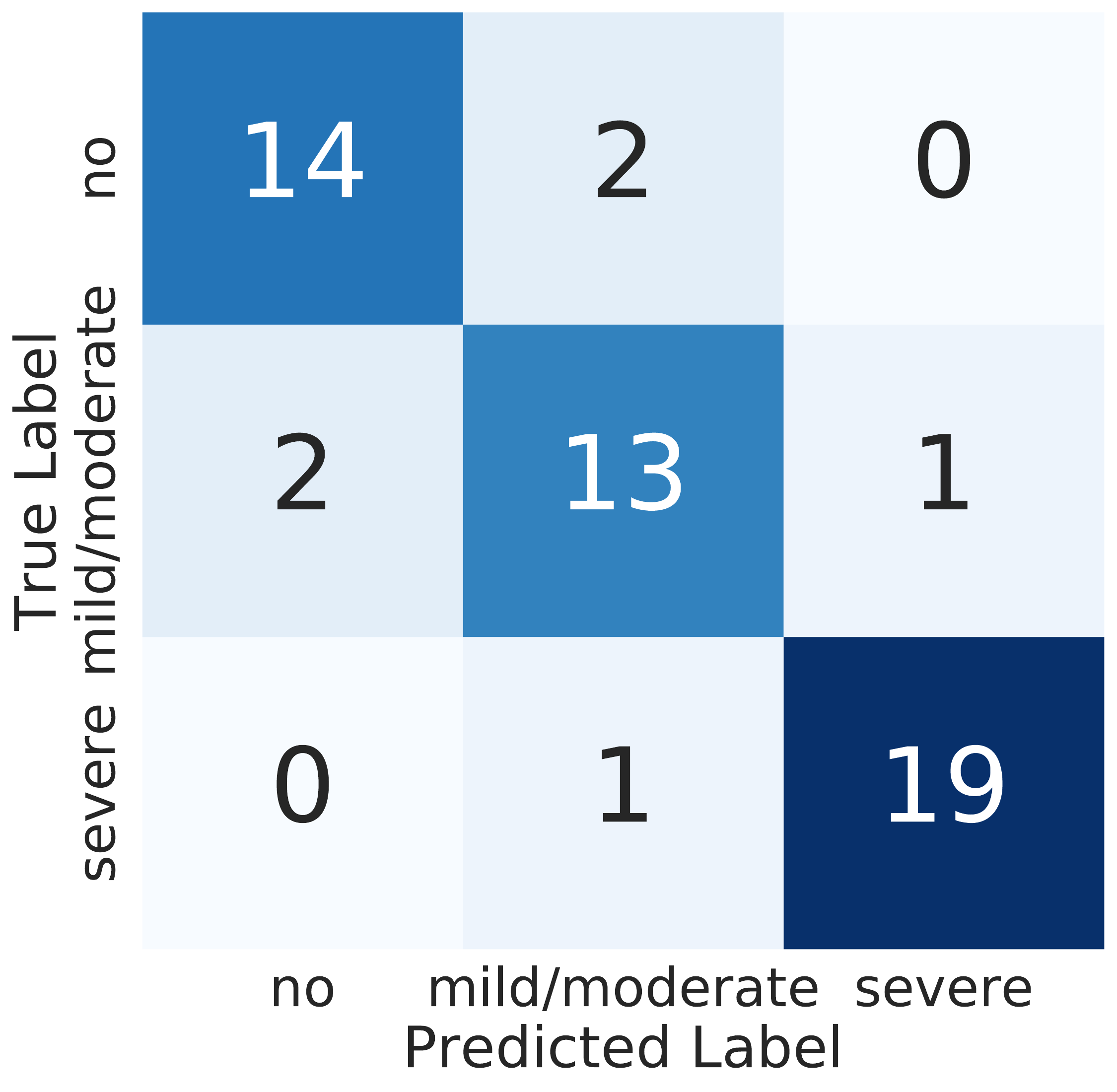}
    &
    \includegraphics[width=\BW\textwidth]{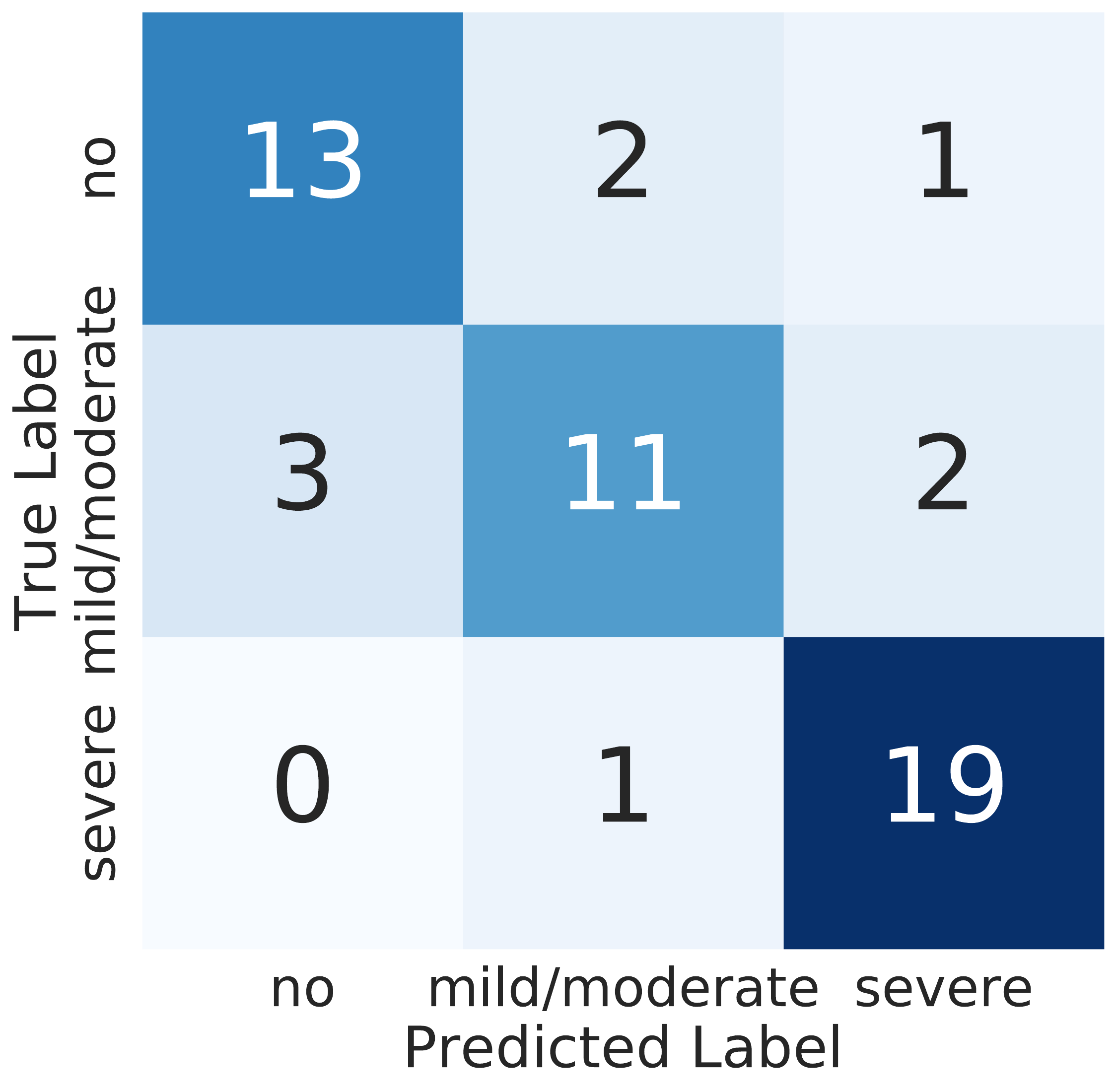}
    &
    \includegraphics[width=\BW\textwidth]{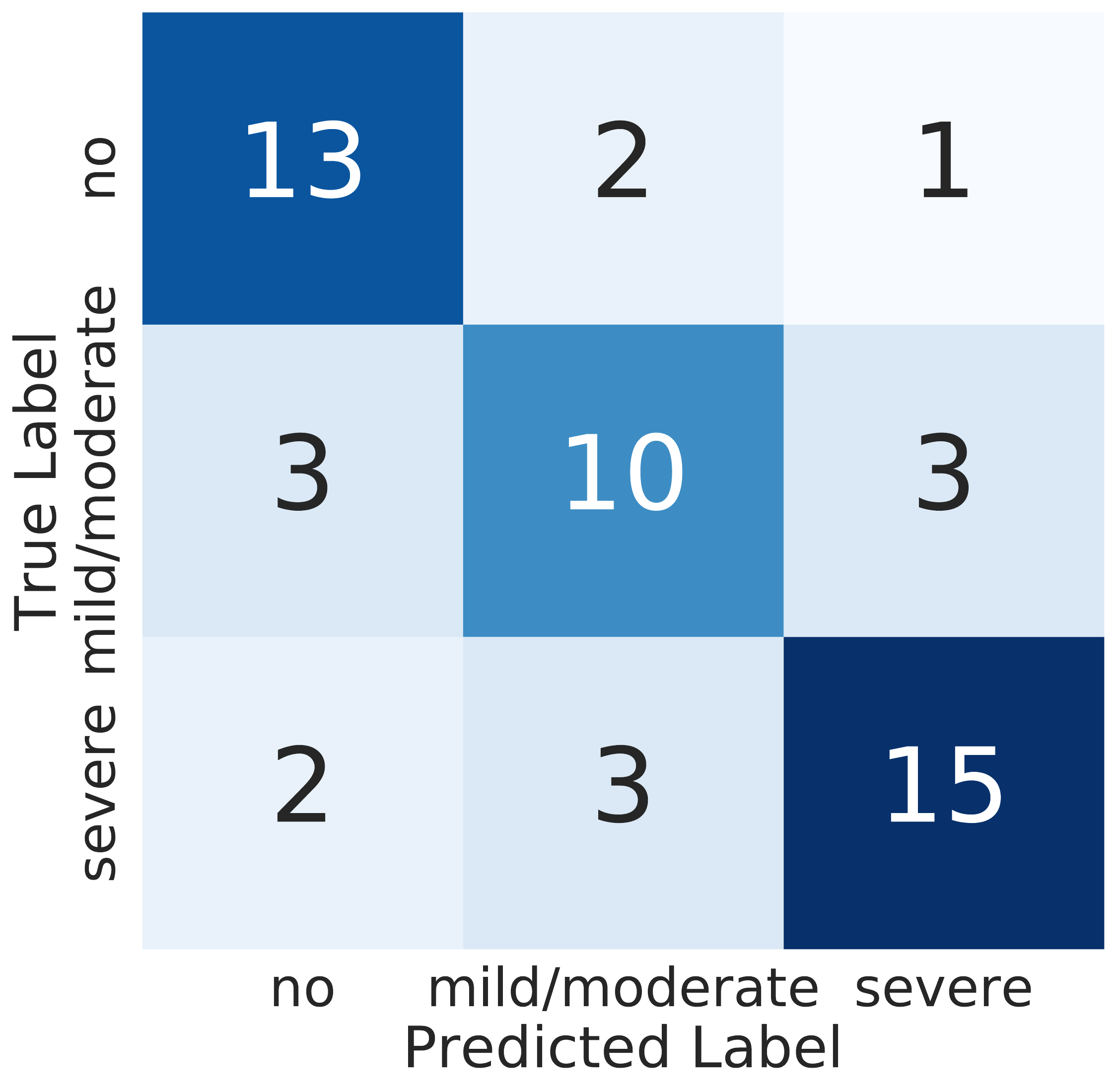}
    &
        \includegraphics[width=\BW\textwidth]{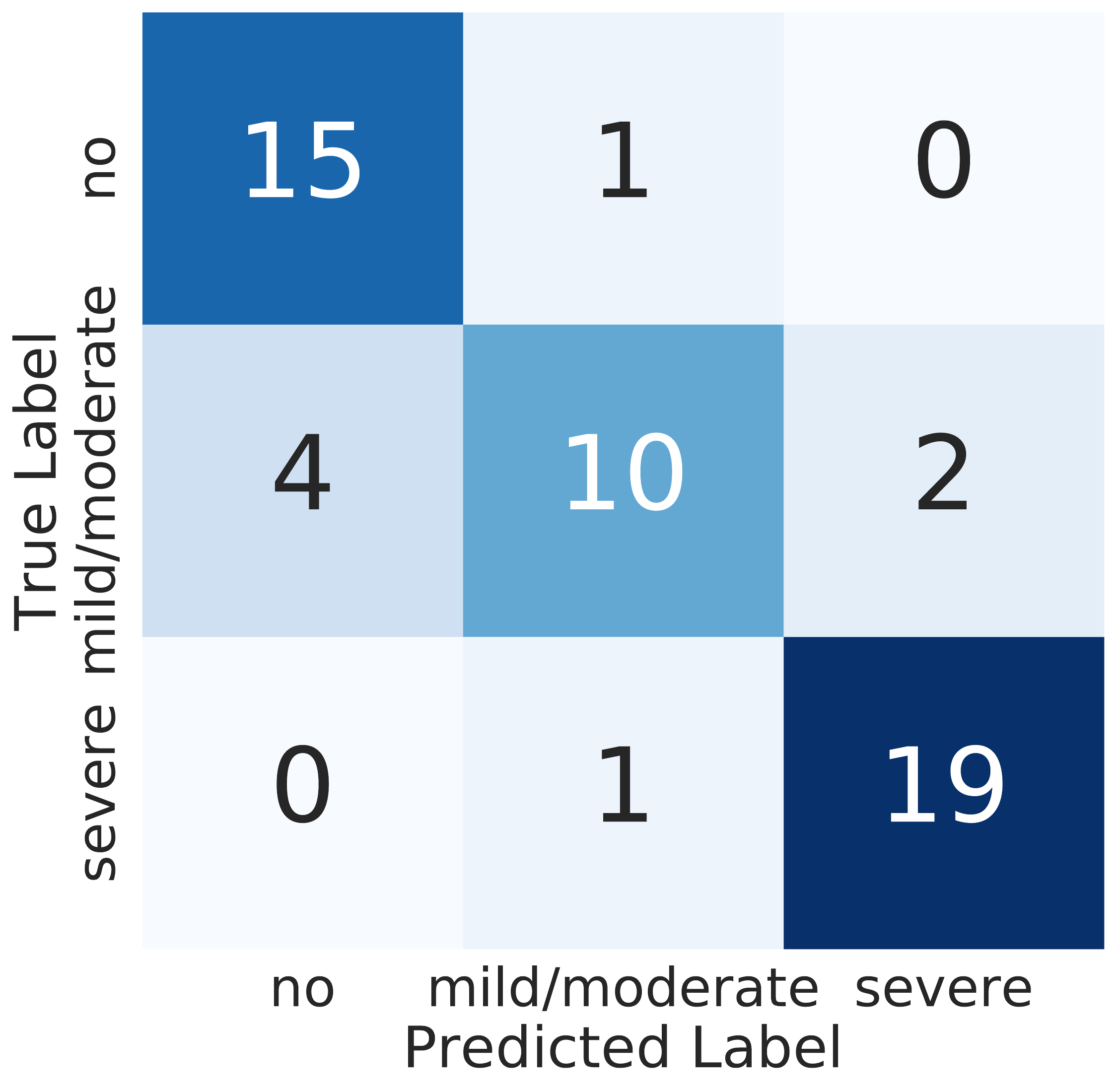}
    \\
    {\rotatebox{90}{~~pretrained MixMatch}}
    & 
    \includegraphics[width=\BW\textwidth]{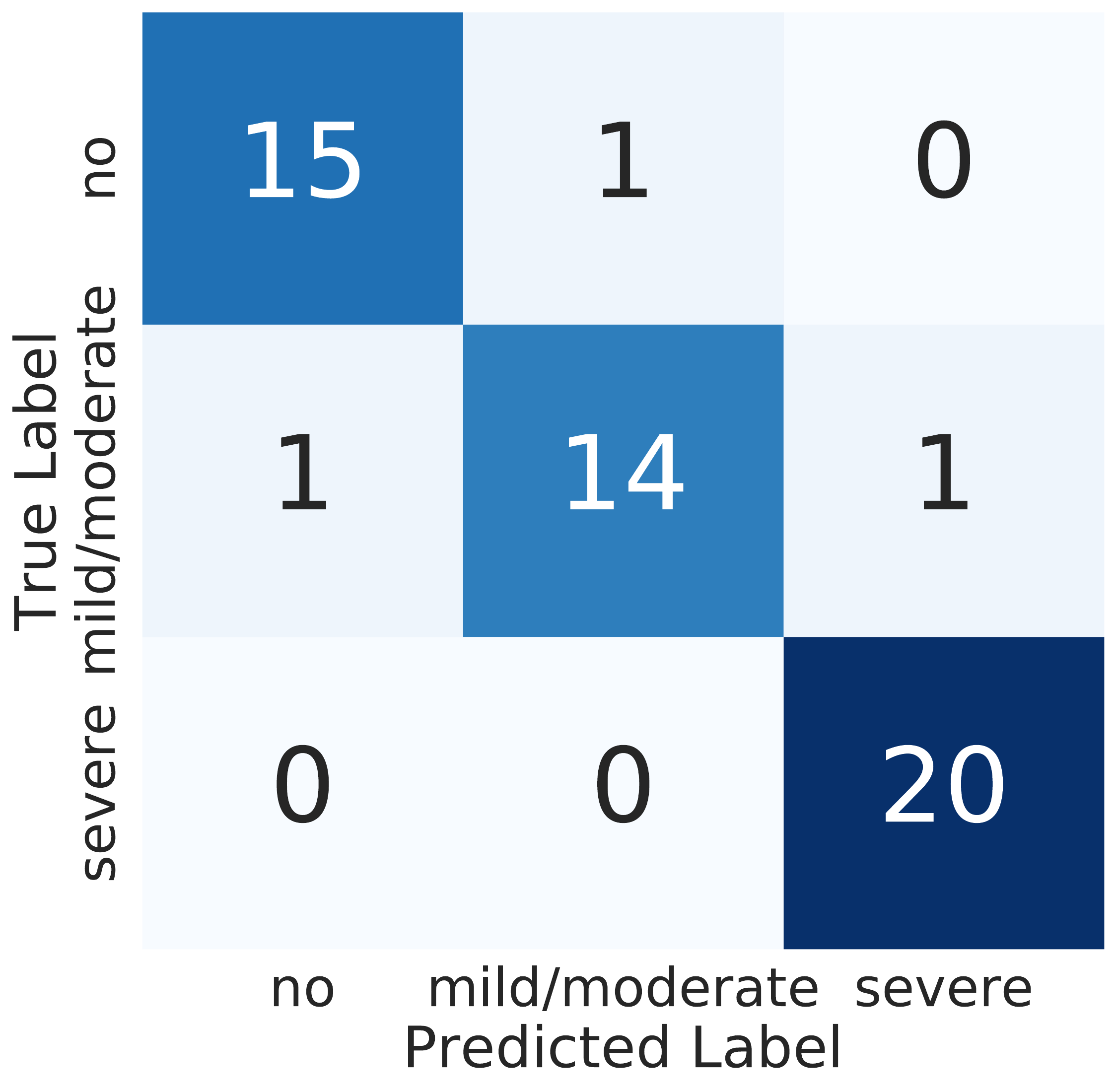}
    &
    \includegraphics[width=\BW\textwidth]{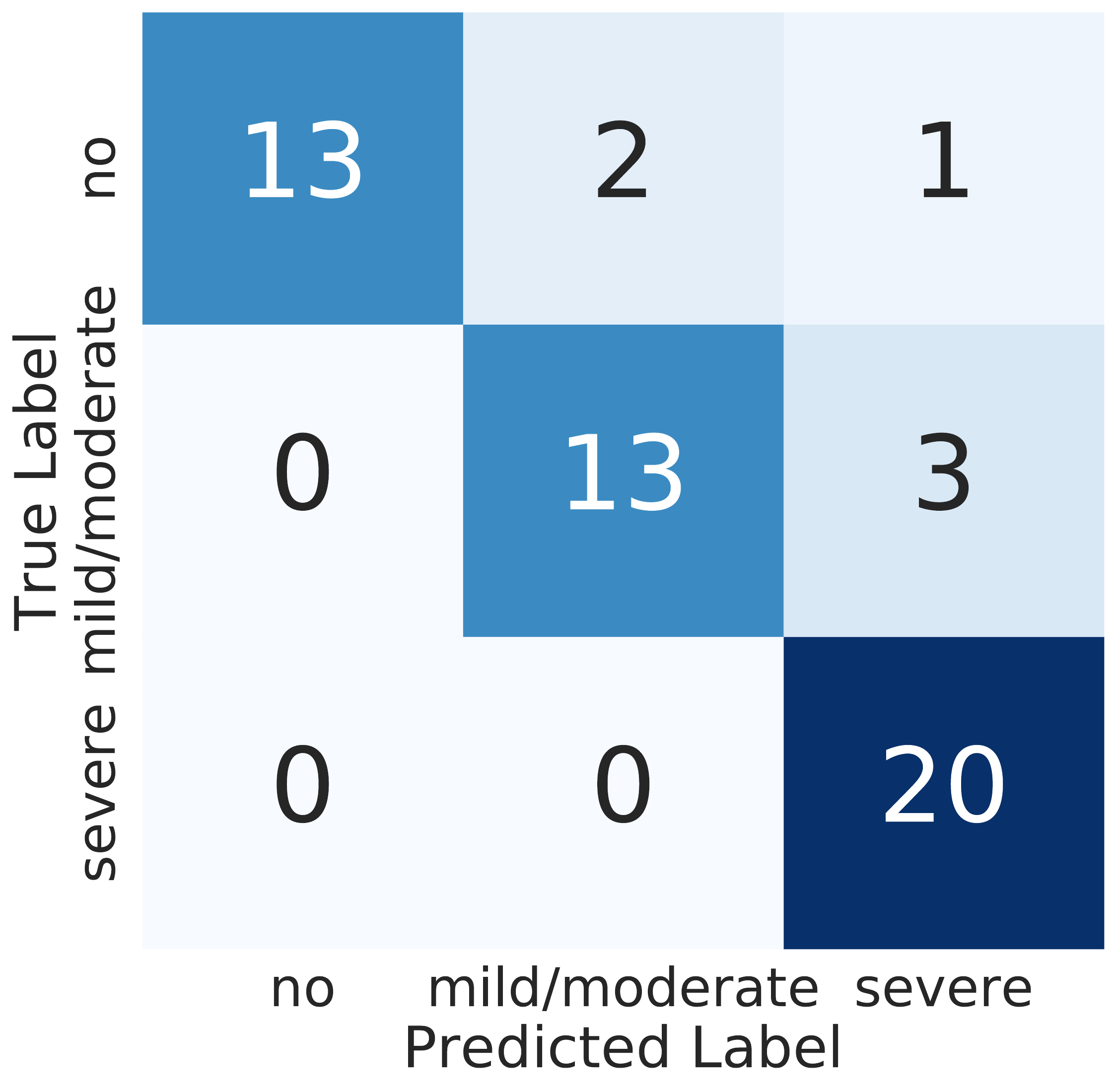}
    &
    \includegraphics[width=\BW\textwidth]{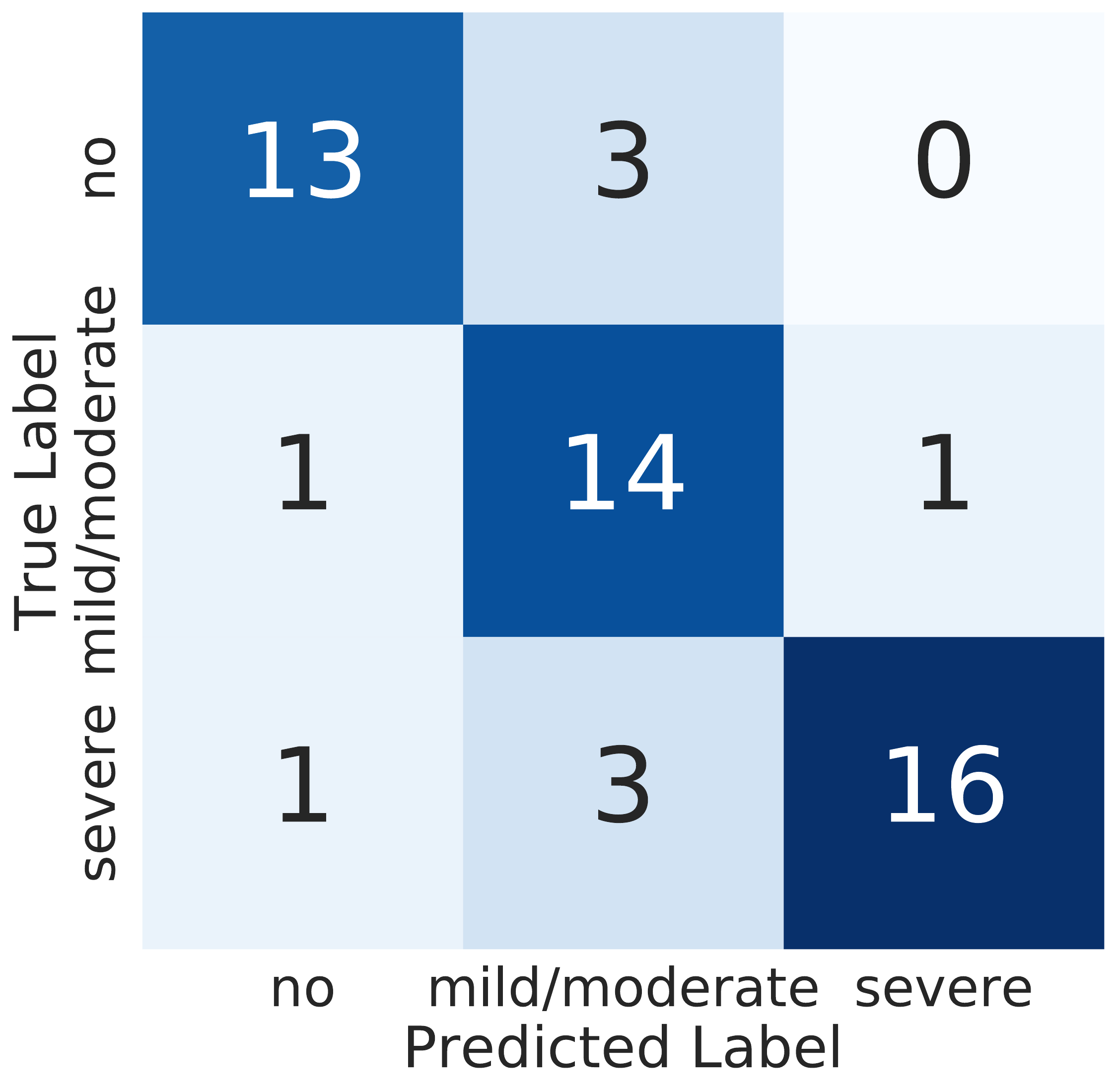}
    &
    \includegraphics[width=\BW\textwidth]{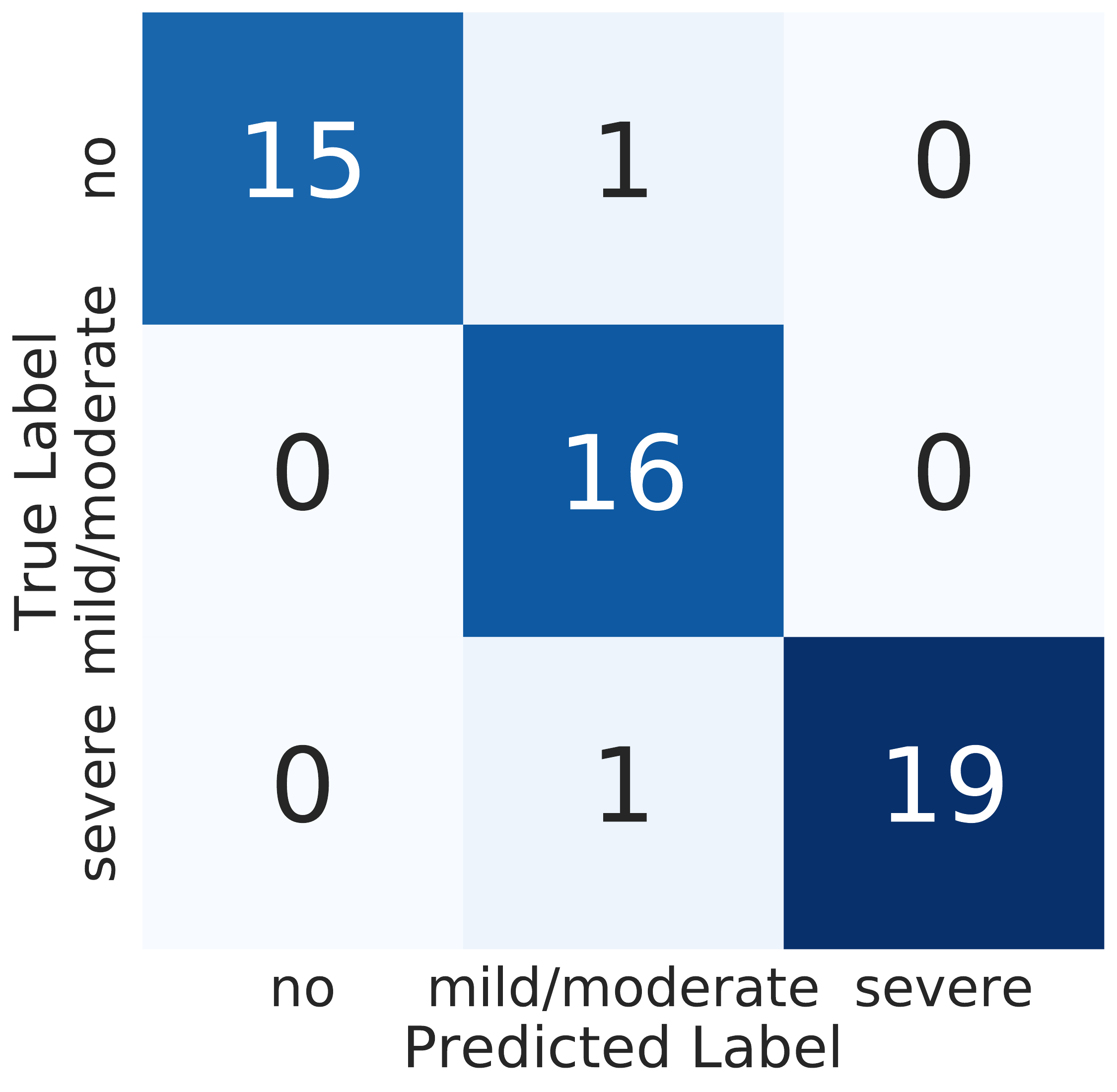}
    \end{tabular}	
    \caption{Confusion matrices for the patient-level AS diagnosis classification, across all four train/test splits of the \textbf{full-size \datasetName-156-52} dataset.
    Each split's test set contains 52 patients.
    We show the labeled-set only basic WRN (top) as well as the best SSL method (bottom).
    Our method improves accuracy compared to the baseline, and most remaining mistakes confuse nearby classes (e.g. ``no AS'' vs. ``mild/moderate AS'') instead of distant classes (e.g. ``no AS'' vs. ``severe AS''). }
    \label{fig:confusion_matrix}
\end{figure}

%% file: discussion.tex
We have developed and evaluated a semi-supervised learning pipeline that can leverage abundant unlabeled data to deliver competitive patient-level diagnostic predictions for the fully-automated preliminary assessment of aortic stenosis. These methods overcome two challenges. First, echocardiograms are challenging to label so the amount of labeled training data is limited. Second, a patient's record will contain hundreds of images, many of which are not relevant for diagnosing AS.
Here, we briefly review the limitations and advantages of our approach.

\paragraph{Limitations.} 
The most important caveat to this work is the need for further independent validation of our methodology.
For logistical reasons, all our data come from one institution.
A detailed evaluation at another institution would be needed to properly assess our proposed pipeline's utility in a prospective setting when used with different patient populations, imaging devices, and sonographers.

A critical and well-known issue with interpretation of echocardiograms is \emph{inter-rater reliability}~\citep{sacchiDopplerAssessmentAortic2018}.
In particular, the distinction between mild and moderate or moderate and severe diagnostic levels can vary across annotators.
All labels in our dataset come from less than 5 expert annotators at one institution.
Further study is needed to understand if our approach could match the consensus of a broader population of annotators.

Several other opportunities to improve our pipeline exist.
Our image processing approach prioritizes simplicity but does not take advantage of recent larger CNN architectures or region-specific attention or segmentation as in some past work on cardiac imaging~\citep{chenDeepLearningCardiac2020}. 
We could use higher-resolution images.
We could include other easily-measured covariates (besides imaging) in our diagnostic model, such as age, demographics, comorbidities, and other cardio-mechanical signals.

A final limitation is that further effort to qualitatively understand what visual signals are driving predictions is needed to build trust.
We plan to investigate saliency maps~\citep{simonyanDeepConvolutionalNetworks2014,selvarajuGradCAMVisualExplanations2020}, though we will be mindful of the limitations of these methods~\citep{adebayoSanityChecksSaliency2018}.
Qualitative insight is key, because fundamentally, MixMatch works by \emph{interpolating} images. It is surprising to us that MixMatch delivered consistently improved results (replicated across several train/test splits) in a real medical imaging scenario, because interpolated echocardiogram images have questionable meaning to human experts.

\paragraph{Advantages.} For potentially fatal conditions like AS, echocardiograms remain the gold standard source of information to produce a diagnosis. Our approach can already reach performance levels (90\% balanced accuracy) that might be useful in a deployed setting (naturally, these must first be reliably replicated in a prospective setting on an external cohort). 
Automatic diagnostic classification pipelines have the potential to identify individuals who would benefit from further screening who otherwise would not be discovered due to limited access to expert cardiologists.

A key aspect of our approach is demonstrating the value of \emph{semi-supervised} learning for a real medical task with class-imbalance (for our view task over 80\% of the images are ``Other'').
Our dataset also includes truly unlabeled data from over 2400 patients, which represents a more authentic test of SSL than previous benchmark datasets.
Overall, our work motivates modern SSL as a promising cost-effective way to improve performance if unlabeled data is abundant, even for real clinical images with substantial diversity.
Especially if labeled sets are small, the gains from SSL may be even greater (see Table~\ref{tab:view classification small}).

A final advantage of our work is the demonstration that patient-level diagnosis benefits from \emph{prioritizing relevant views}. Building on \citet{madaniDeepEchocardiographyDataefficient2018}, who showed promising SSL diagnosis given manually-curated views, our SSL methods can deliver effective diagnoses given an uncurated set of \emph{all} available images, even when most depict irrelevant views.

We hope our study marks a step toward effective early detection of aortic stenosis that can enable helpful interventions. We further hope this study and the accompanying dataset release offer a reproducible template for improving patient outcomes for other diseases where medical imaging is key and labeled data is scarce.

%% file: acks.tex
\section*{Acknowledgements}

All authors gratefully acknowledge financial support from
the Pilot Studies Program at the Tufts Clinical and Translational Science Institute
(Tufts CTSI NIH CTSA UL1TR002544).
Author BW is supported by K23AG055667 (NIH-NIA). 
Authors HZ and MCH thank the Office of the Vice Provost for Research at Tufts University for support for this project under a ``Tufts Springboard'' award.

%% file: Appendix.tex
\newpage
\section{Dataset Visualizations}
\label{sec:app_dataset_visuals}

\subsection{Examples of each view class}
\newcommand{\BC}{0.33}
\setlength{\tabcolsep}{0.1cm}
\begin{figure}[!h]
\begin{tabular}{c c c c}
    PLAX  & PSAX & OTHER 
    \\
    \includegraphics[width=\BC\textwidth]{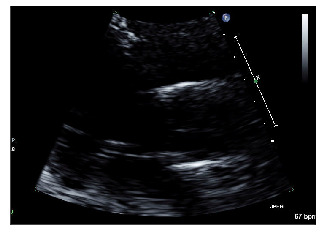}
    &
    \includegraphics[width=\BC\textwidth]{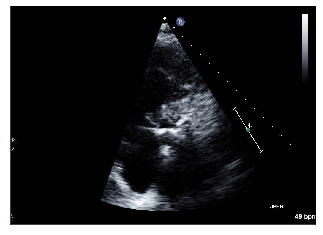}
    &
    \includegraphics[width=\BC\textwidth]{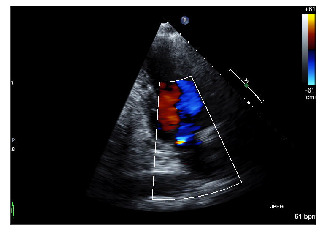}
    &
   
    \\
    
    \includegraphics[width=\BC\textwidth]{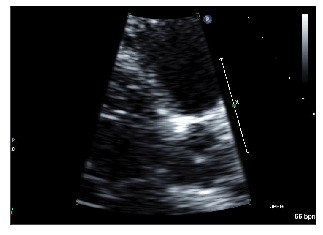}
    &
    \includegraphics[width=\BC\textwidth]{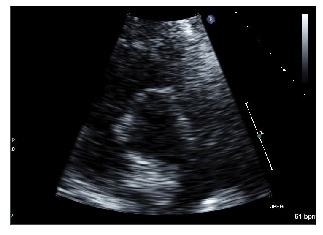}
    &
    \includegraphics[width=\BC\textwidth]{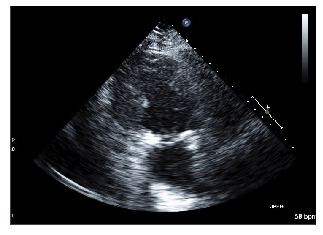}
    &
   
     \\
     
     \includegraphics[width=\BC\textwidth]{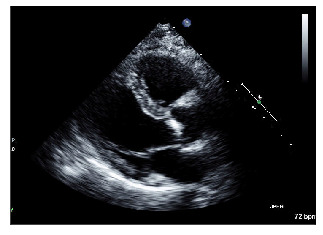}
    &
    \includegraphics[width=\BC\textwidth]{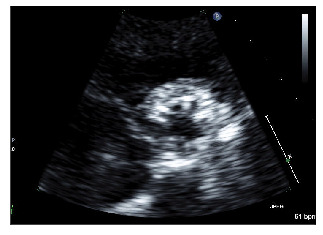}
    &
    \includegraphics[width=\BC\textwidth]{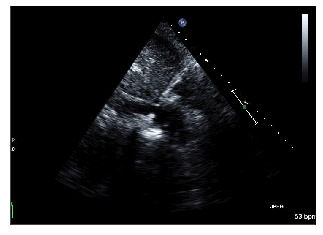}
    &
   
     \\
     
     \includegraphics[width=\BC\textwidth]{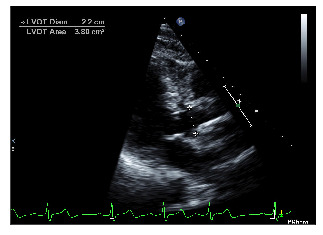}
    &
    \includegraphics[width=\BC\textwidth]{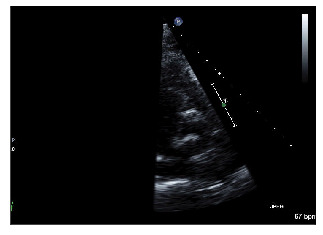}
    &
    \includegraphics[width=\BC\textwidth]{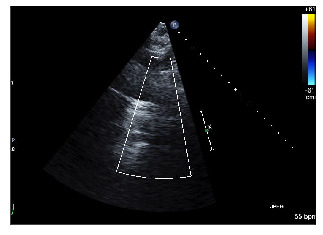}
    &
   
    \end{tabular}	
    \caption{Examples of images for each possible view label in our dataset. \emph{From left to right:} Four examples of peristernal long axis (PLAX) view, four examples of peristernal short axis (PSAX) view, and four examples of other kinds of view in our ``Other'' class. }
    \label{fig:VIEW_SAMPLES_APPENDIX}
\end{figure}

\newpage
\subsection{Examples of each view for a Severe AS patient}
\newcommand{\BA}{0.33}
\setlength{\tabcolsep}{0.1cm}
\begin{figure}[!h]
\begin{tabular}{c c c c}
    PLAX  & PSAX & OTHER 
    \\
    \includegraphics[width=\BA\textwidth]{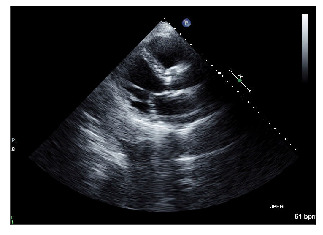}
    &
    \includegraphics[width=\BA\textwidth]{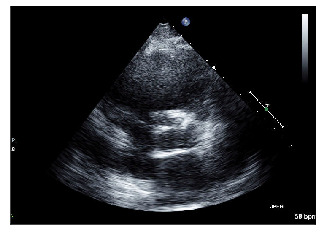}
    &
    \includegraphics[width=\BA\textwidth]{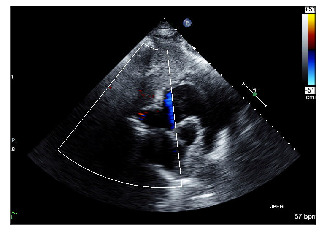}
    &
    
    \\
    
    \includegraphics[width=\BA\textwidth]{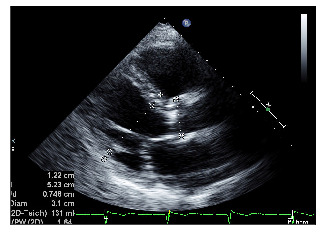}
    &
    \includegraphics[width=\BA\textwidth]{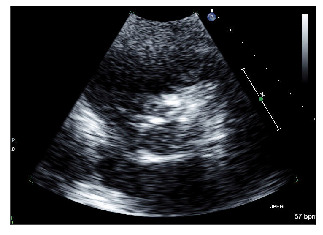}
    &
    \includegraphics[width=\BA\textwidth]{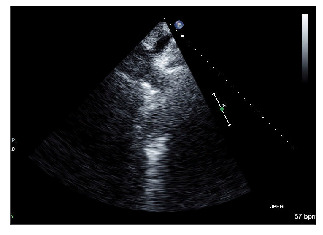}
    &
   
     \\
     
     \includegraphics[width=\BA\textwidth]{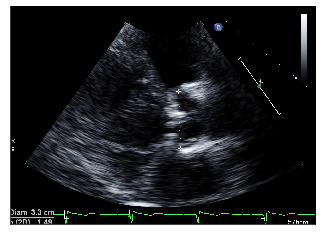}
    &
    \includegraphics[width=\BA\textwidth]{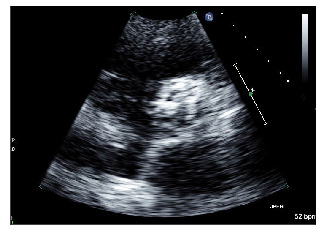}
    &
    \includegraphics[width=\BA\textwidth]{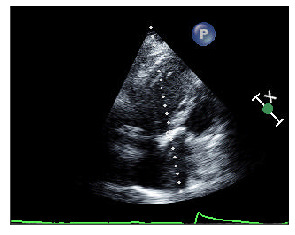}
    &
  
    \end{tabular}	
    \caption{Examples of images from a patient with Severe AS in our dataset. \emph{From left to right:} Three examples of parasternal long axis (PLAX) view, three examples of parasternal short axis (PSAX) view, and three examples of other kinds of view in our ``Other'' class. }
    \label{fig:PatientSevereAS}
\end{figure}

\newpage
\subsection{Examples of each view for a No AS patient}
\newcommand{\BB}{0.33}
\setlength{\tabcolsep}{0.1cm}
\begin{figure}[!h]
\begin{tabular}{c c c c}
    PLAX  & PSAX & OTHER 
    \\
    \includegraphics[width=\BB\textwidth]{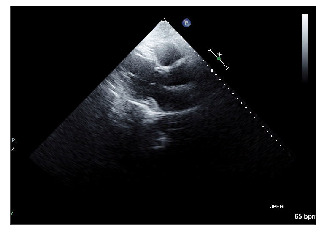}
    &
    \includegraphics[width=\BB\textwidth]{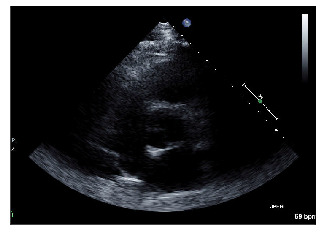}
    &
    \includegraphics[width=\BB\textwidth]{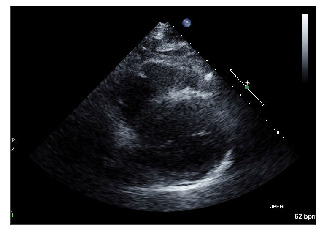}
    &
    
    \\
    
    \includegraphics[width=\BB\textwidth]{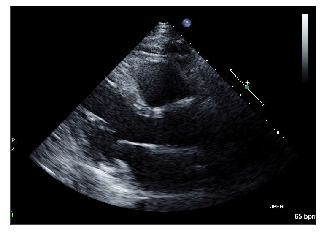}
    &
    \includegraphics[width=\BB\textwidth]{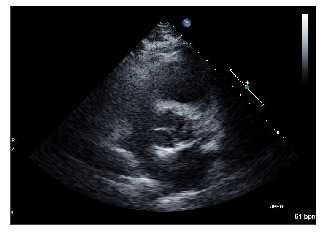}
    &
    \includegraphics[width=\BB\textwidth]{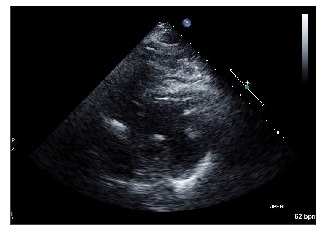}
    &
   
     \\
     
     \includegraphics[width=\BB\textwidth]{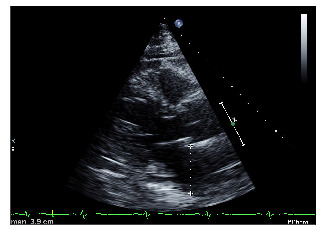}
    &
    \includegraphics[width=\BB\textwidth]{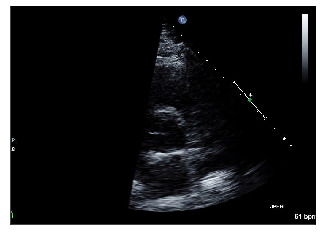}
    &
    \includegraphics[width=\BB\textwidth]{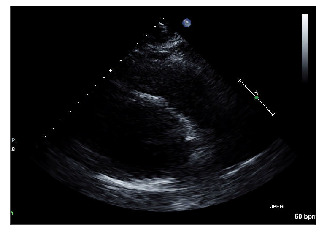}
    &
  
    \end{tabular}	
    \caption{Examples of images from a patient with No AS in our dataset. \emph{From left to right:} Three examples of parasternal long axis (PLAX) view, three examples of parasternal short axis (PSAX) view, and three examples of other kinds of view in our ``Other'' class. }
    \label{fig:PatientNoAS}
\end{figure}

\newpage 
\section{Further Results}

\subsection{Assessment of ensembling}

Table~\ref{tab:best_single_checkpoint_VS_ensemble_FS_echo260} compares using a single checkpoint (one point estimate of neural network weight vector $\theta$) to using an ensemble of parameters aggregated from the last 25 checkpoints (one per epoch).

\begin{table}[!h]
    \centering
    \begin{tabular}{c|cccc|c}
    \textit{Diagnosis classification} & Split 1  & Split 2 & Split 3 & Split 4 & Average\\
    \hline
    Best single checkpoint  & 61.81 & 59.79 & 56.05 & 64.21 & 60.46\\
    Ensemble  & 62.95 & 61.03 & 56.58 & 63.84 & \textbf{61.13}
	\\ \hline
    \textit{View classification}  &   &  &  &  & 
    \\ \hline
    Best single checkpoint  & 93.03 & 93.24 & 92.39 & 93.79 & 93.11\\
    Ensemble  & 92.37 & 93.24 & 93.72 & 93.87 & \textbf{93.30}\\
    \end{tabular}
    \caption{Comparing best single checkpoint performance with ensemble performance on \textbf{Full-size \datasetName-156-52}}
    \label{tab:best_single_checkpoint_VS_ensemble_FS_echo260}
\end{table}

\subsection{Patient-level diagnosis performance on bonus heldout set}

Table~\ref{tab:diagnosis classification patient unlabeled_heldout_174} examines the performance of the best labeled-set-only methods and MixMatch methods on the 174 patient studies that have diagnosis but no view labels.
 While the images used here were originally included in the unlabeled training set (which was used to train SSL methods like MixMatch), the diagnosis labels were not provided at all during training time. 
 We thus still believe this is an authentic test of generalization given the scarcity of labeled data available for our task.
 Of course, additional independent evaluation (especially from another institution) is needed.

\begin{table}[!h]
    \centering
    \begin{tabular}{l l l|rrrr|c}
    Pretrain & Method & Voting
    & Split 1  & Split 2 & Split 3 & Split 4 & average\\
    \hline
    & Basic WRN & Simple average & 76.73 & 75.25 & 76.87 & 81.88 & 77.68\\
    & Basic WRN & View-prioritized & 73.63 & 83.21 & 79.70 & 80.08 & 79.18\\
    \hline
    & MixMatch & Simple average & 85.32 & 76.29 & 74.14 & 79.95 & 78.93\\
    view & MixMatch & Simple average & 83.36 & 77.96 & 75.61 & 81.37 & 79.58\\
    & MixMatch & View-prioritized & 83.27 & 83.76 & 82.34 & 82.83 & \textbf{83.05}\\
    view & MixMatch & View-prioritized & 82.53 & 86.15 & 79.62 & 83.27 & 82.89\\
    \end{tabular}
    \caption{Patient-level AS Severity Diagnosis Classification on the \textbf{bonus heldout set} of 174 patients for whom we have diagnosis labels only (no view labels). We show balanced accuracy on models trained on each of the four folds on four \textbf{full-size \datasetName-156-52} dataset.
    }
    \label{tab:diagnosis classification patient unlabeled_heldout_174}
\end{table}

\subsection{Assessment of MixMatch hyperparameter sensitivity}

In Table~\ref{tab:MixMatch hyperparameters ablation study}, we consider four possible strategies for setting the hyperparameters of MixMatch, varying two  key settings for the weight on unlabeled loss $\lambda$. First, we vary whether the final value of $\lambda$ is set to its \emph{best} value among a grid of candidates (based on validation set performance), or \emph{fixed} to a constant.
Second, we vary whether $\lambda$ remains fixed over iterations throughout a training run, or is updated over iterations on a linear ramp schedule from 0 to its final target value. 

From this comparison, we see we consistent gains across splits (average gain across splits of over 1.6\% balanced accuracy) for using a delayed ramp up schedule with target value selected via grid search.

\begin{table}[!h]
    \centering
    \begin{tabular}{l l| rrrr | r}
    Final $\lambda$ value & $\lambda$ update schedule & Split 1  & Split 2 & Split 3 & Split 4 & Average\\
    \hline
    best on val & Delayed ramp-up  & 65.57 & 62.69 & 60.87 & 66.29 & 63.86\\
    best on val & Immediate ramp-up & 65.07 & 61.87 & 60.82 & 65.37 & 63.28\\
    best on val & Constant  & 65.03 & 61.52 & 58.87 & 65.22 & 62.66\\
    100 (fixed) & Constant & 63.94 & 61.79 & 58.87 & 64.35 & 62.24\\
    \end{tabular}
    \caption{Ablation study of different settings of the unlabeled loss weight $\lambda$ for MixMatch. AS severity diagnosis classification for individual images on the \textbf{full-size \datasetName-156-52} dataset. showing balanced accuracy averaged over the test sets from multiple folds (each fold’s test set contains all images from 52 patients). }
    \label{tab:MixMatch hyperparameters ablation study}
\end{table}

\subsection{Assessment of alternative view prioritization strategy using thresholding}

An anonymous reviewer suggested an alternative strategy for prioritizing images of relevant view.
The alternative strategy works as follows: for each image, we compute the predicted probability that the image is a ``relevant view'' (either PLAX and PSAX) by summing the probabilities of each view type.
However, instead of using this raw probability as a weight (as our chosen method does), we use a \emph{cutoff threshold} and simply average the diagnosis predictions of images whose relevant view probability is above the cutoff.
For each patient, we use the majority vote prediction of the diagnosis from the images of relevant views.
The value of the cutoff threshold is selected using the validation set to maximize balanced accuracy.

Table~\ref{tab:Suggested_Aggregation_Ablation} shows the performance of this strategy (``threshold-then-average'') on the full-size dataset.
Using this alternative prioritization strategy together with our suggested methodology for patient-level diagnosis (using MixMatch, pretraining on view), we find the average test set balanced accuracy is around 85.8\%, while the weighted average strategy in the main paper achieves over 90\% balanced accuracy. We take this as reasonably decisive evidence that a weighted average (rather than a simple cutoff) should be preferred.

\begin{table}[!h]
    \centering
    \begin{tabular}{l l l|rrrr|c}
    Pretrain & Method & Aggregation across images
    & Split 1  & Split 2 & Split 3 & Split 4 & average\\
    \hline
    & Basic WRN & Threshold-then-Average & 85.42 & 86.25 & 79.17 & 92.50 & 85.84 \\
    & MixMatch & Threshold-then-Average & 83.33 & 84.17 & 77.50 & 94.58 & 84.90 \\
    view & MixMatch & Threshold-then-Averagen & 86.67 & 80.00 & 82.50 & 94.17 & 85.84\\
    \end{tabular}
    \caption{Alternative view-prioritizing strategy for patient-level AS severity diagnosis classification on the \textbf{full-size \datasetName-156-52} dataset, showing balanced accuracy on the test set across multiple folds (each fold’s test set contains 52 patients).}
    \label{tab:Suggested_Aggregation_Ablation}
\end{table}

\subsection{ROC Curve of patient-level diagnosis: no AS vs. mild/moderate/severe AS}

\begin{figure}[!h]
\begin{tabular}{c c}
	\includegraphics[width=0.43\textwidth]{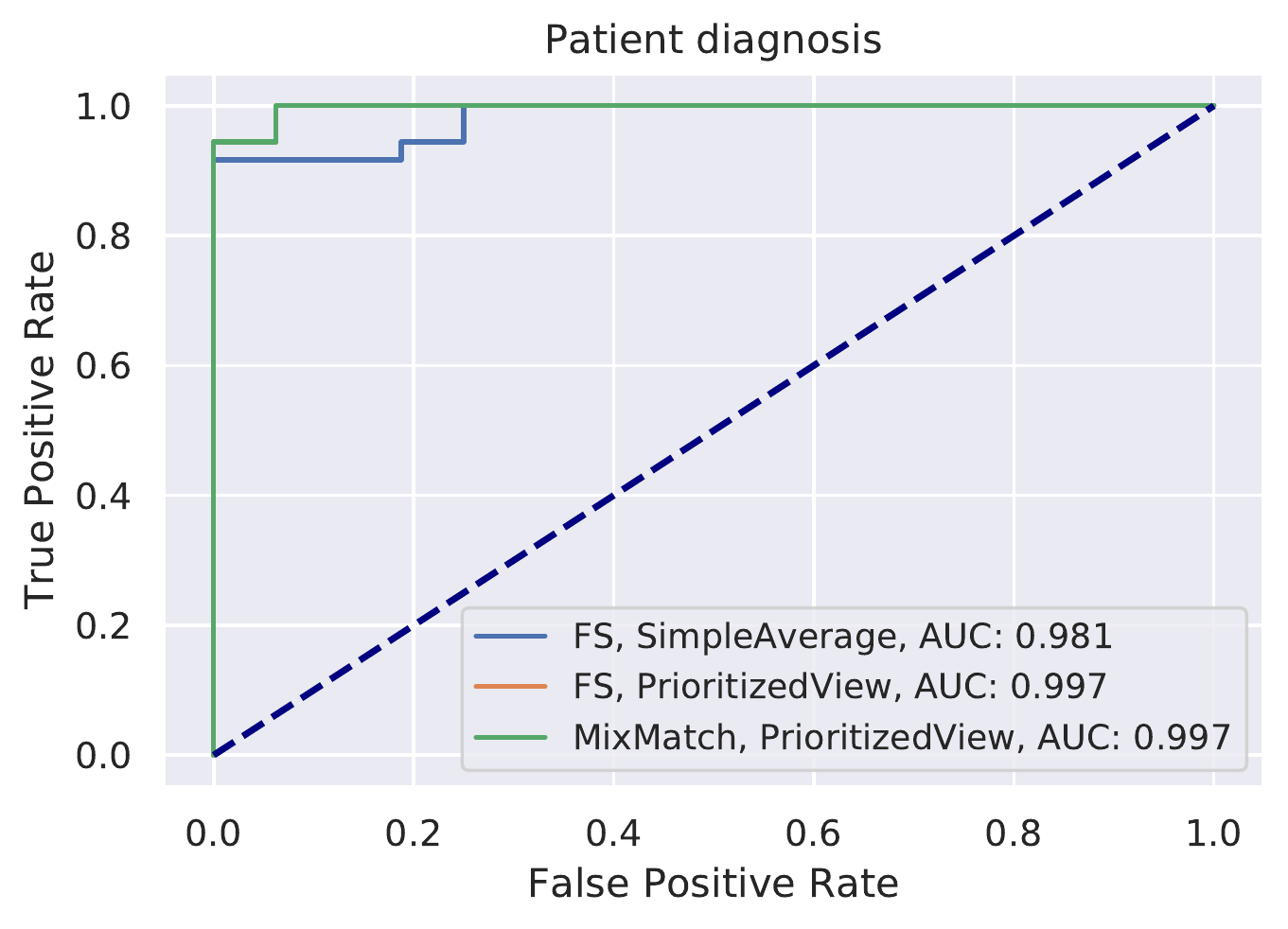}
	&
    \includegraphics[width=0.43\textwidth]{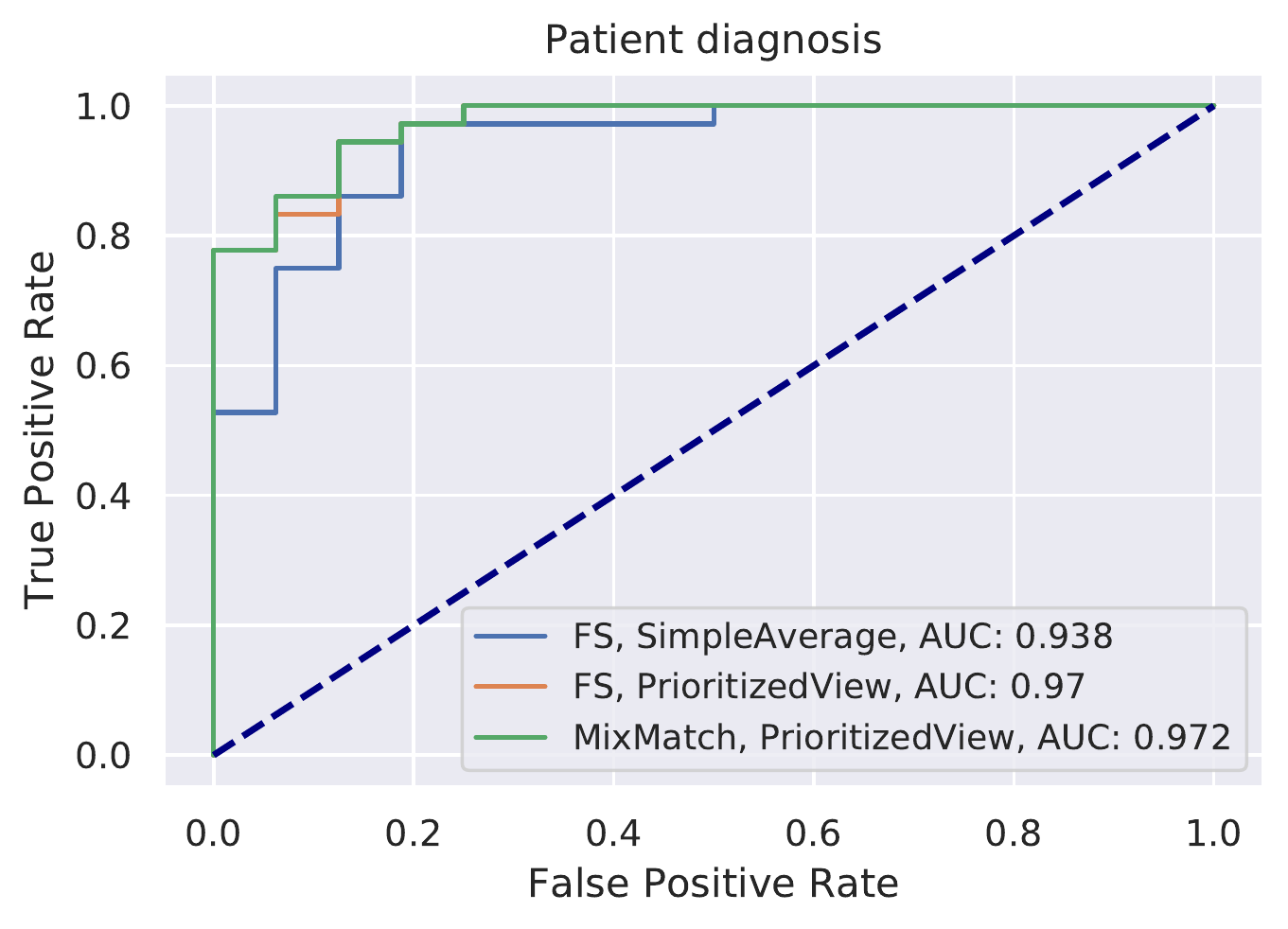}
	\\
	(a) Split 1 & (b) Split 2
	\\
	\includegraphics[width=0.43\textwidth]{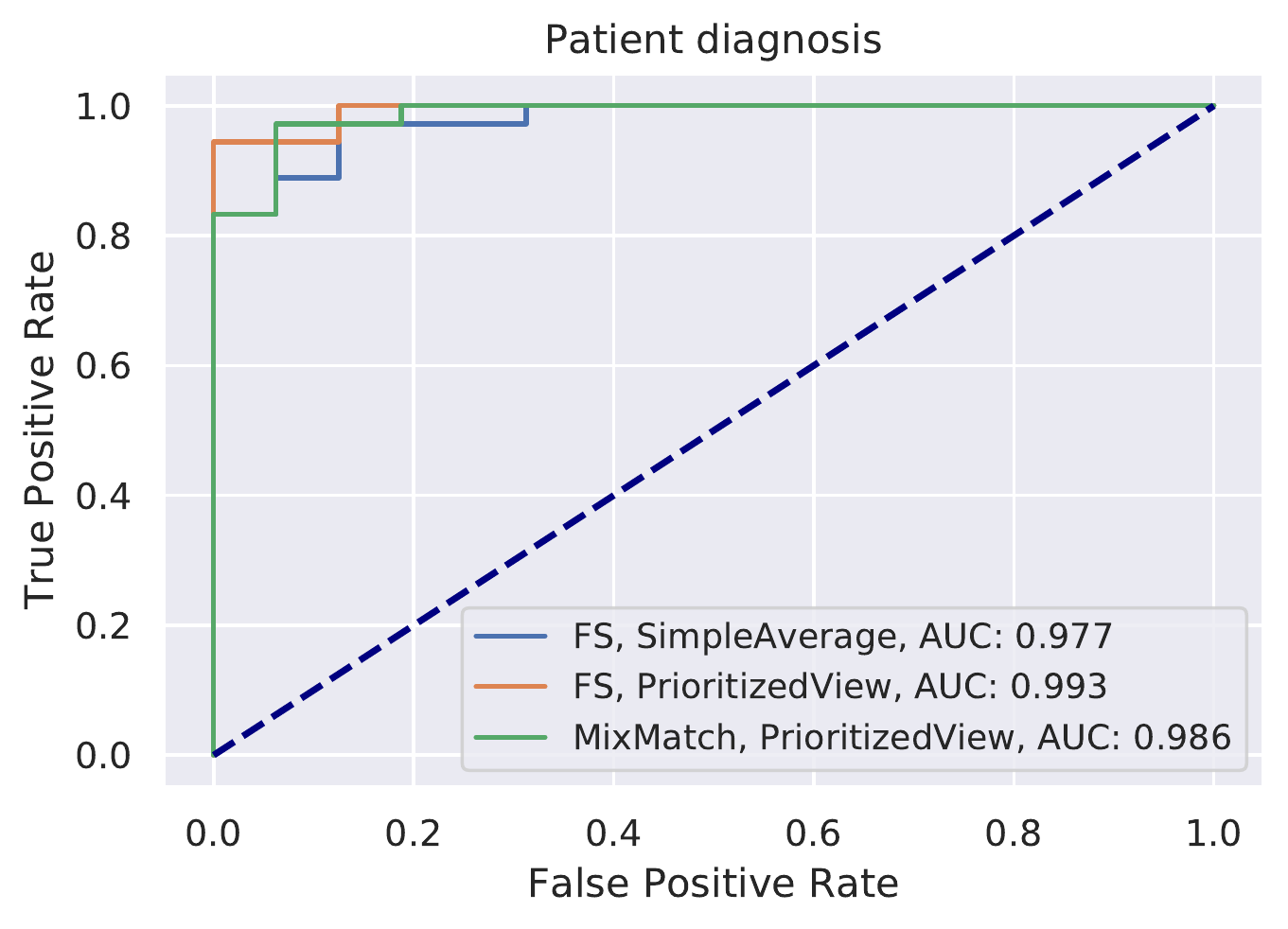}
	&
    \includegraphics[width=0.43\textwidth]{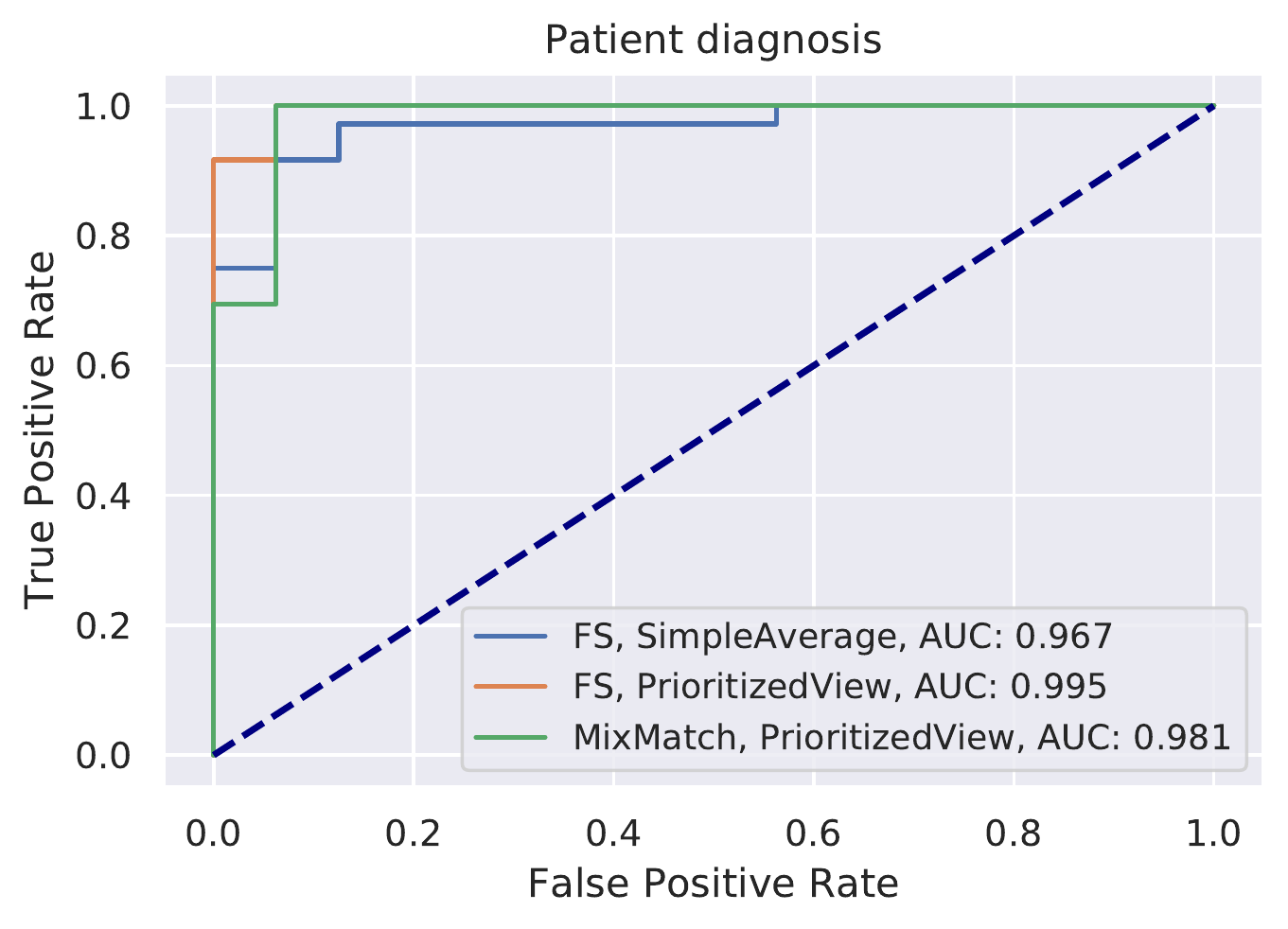}
	\\
	(c) Split 3 & (d) Split 4
\end{tabular}
    
\caption{ROC curves for binary diagnosis task (no AS vs ``mild/moderate/severe AS'') on \textbf{full-size \datasetName-156-52}.
    }
    \label{fig: No AS vs Some AS}
\end{figure}

Fig.~\ref{fig: No AS vs Some AS} shows receiver operating curves for several methods for the task of distinguishing no AS vs Some AS (which aggregates both the mild/moderate and severe levels in the 3-level diagnosis task of the main paper).

\section{Methodological Details}

\subsection{Image processing details}
\label{sec:removing_doppler}

\paragraph{Removing doppler images.}
In the raw data of all imagery available for an echocardiogram study, 
we obtained TIFF files that represent both cineloops and Doppler images.

We verified in our labeled set that all Doppler images have one of the following landscape aspect ratio $(831, 323)$, $(901, 384)$, $(901, 390)$, $(704, 305)$, $(831, 421)$, $(901, 469)$ or $(563, 294)$. Only the Dopplers have these aspect ratios. We thus filtered out Doppler completely via these aspect ratios. 

\paragraph{Downsizing}
The original images are provided as high-resolution TIFF format images (hundreds of pixels per side) of varying aspect ratios. Generally, we can expect that both view and diagnosis classifiers would perform better given higher-resolution input (and holding other factors the same). The main trade-off of processing higher-resolution images is increased runtime and memory requirements. In our preliminary experiments, we compared downsizing all images to a standard square aspect ratio at 3 possible sizes: 32x32, 64x64 and 128x128. We found that 64x64 achieves a good balance between model performance and computation cost. 
A prior study by \citet{madaniDeepEchocardiographyDataefficient2018} provides a more extensive study of optimal resolution size. The interested reader can refer to their work for more details.

\subsection{Architecture Settings and Hyperparameters}
\label{sec:arch_and_hyperparameters}

\paragraph{Weighted cross-entropy for labeled loss}
To counteract the effect of class imbalance in the dataset, we use weighted cross-entropy for the labeled loss. For an input image $x$ whose true label $y$ indicates it belongs to class $c$, the weighted cross-entropy assumes the following form:
\begin{align}
\mathcal{L}^L(\theta, x) = - w_{c} \log \hat{p}_{c}(\theta, x),
\end{align}
where $\hat{p}_{c}$ is the predicted probability of class $c$. The weight $w_{c}$ is calculated using the training set statistics as follow:
\begin{align}
w_{c} = \frac{\prod_{k\neq c}{N_{k}}}{\sum_{j}\prod_{k \neq j}{N_{k}}}
\end{align}
where $N_{k}$ is the number of images of class $k$ in the training set.

\paragraph{Common architecture.}
Following~\citet{oliverRealisticEvaluationDeep2018}, for all considered methods, we use the \emph{same} backbone neural network architecture: a wide residual network~\citep{zagoruykoWideResidualNetworks2017} with 28 layers (WRN-28), which has total of 5,931,683 parameters.
This same network architecture is used in the original MixMatch evaluation~\citep{berthelotMixmatchHolisticApproach2019} with promising results.

\paragraph{Common training protocol.}
All SSL methods we consider follow the loss minimization framework with two primary losses (one for ``labeled'' data and one for ``unlabeled'' data) in Eq.~\eqref{eq:standard-SSL-loss-template}.
We allow every method to train for 32 epochs (where each epoch processes $2^{16}$ images, as in \citet{berthelotMixmatchHolisticApproach2019}).
Our preliminary experiments suggest that after 30 epochs all methods effectively converge in terms of validation balanced accuracy. 

\paragraph{Common regularization.}
For all methods, we expect performance will be vulnerable to overfitting, so we impose an L2-norm penalty on the weights $\theta$, also known as weight decay. Each method selects its preferred value of this penalty strength hyperparameter. We searched values in [0.0002, 0.002, 0.02].

\paragraph{Common optimization.}
We use ADAM \citep{kingma2014adam} to optimize each model.
Each method selects the value of the step size (learning rate) as a hyperparameter. We experimented with 0.002 and 0.0007

\paragraph{Hyperparameters for Pseudo-Label.}
Beyond the usual hyperparameters for our loss-minimization SSL framework, another important hyperparameter for pseudo-label is the threshold $\tau$. We find that performance is not very sensitive to the chosen $\tau$ value as long as it is within a certain range. We set $\tau$ to 0.95, as done in past literature that evaluates Pseudo-Label as an SSL method ~\citep{oliverRealisticEvaluationDeep2018,berthelotMixmatchHolisticApproach2019, berthelotRemixmatchSemisupervisedLearning2019, sohnFixmatchSimplifyingSemisupervised2020}.

\paragraph{Hyperparameters for VAT.}
Beyond the usual hyperparameters for our SSL framework, for VAT we need to select a value for $\epsilon$.
In \citet{miyatoVirtualAdversarialTraining2019}, the authors claimed that they can achieve superior performance by tuning only $\epsilon$ and fixing $\lambda$ to 1. In our experiment, we used the default $\lambda$ as in \cite{berthelotMixmatchHolisticApproach2019} and searched the value of $\epsilon$ in [2, 6, 18], together with learning rate and weight decay. We select the best hyperparameters using validation set performance.

\paragraph{Hyperparameters for MixMatch.}
Beyond the usual hyperparameters for our SSL framework, the key hyperparameters for MixMatch include the number of augmentations $K$, the temperature $T>0$ used for sharpening, interpolation hyperparameter $\alpha$ and unlabeled loss coefficient $\lambda$. We set $K=2$, $T=0.5$, and $\alpha=0.75$ as done in \citet{berthelotMixmatchHolisticApproach2019}, and search for $\lambda$ in the range [10, 30, 75, 100, 130] using validation set. 

\paragraph{Hyperparameters for Multitask training.}
We searched $\gamma$, the hyperparameter that control the strength of the auxilliary view loss in Eq.~\eqref{eq:multitask}, in the range [10, 3, 1, 0.3, 0.1]. The best $\alpha$ is selected together with other hyperparameters on validation set.

%% file: main.bbl
\begin{thebibliography}{53}
\providecommand{\natexlab}[1]{#1}
\providecommand{\url}[1]{\texttt{#1}}
\expandafter\ifx\csname urlstyle\endcsname\relax
  \providecommand{\doi}[1]{doi: #1}\else
  \providecommand{\doi}{doi: \begingroup \urlstyle{rm}\Url}\fi

\bibitem[Adebayo et~al.(2018)Adebayo, Gilmer, Muelly, Goodfellow, Hardt, and
  Kim]{adebayoSanityChecksSaliency2018}
J.~Adebayo, J.~Gilmer, M.~Muelly, I.~Goodfellow, M.~Hardt, and B.~Kim.
\newblock Sanity {{Checks}} for {{Saliency Maps}}.
\newblock In \emph{Advances in {{Neural Information Processing Systems}}},
  2018.
\newblock
  \url{https://papers.neurips.cc/paper/2018/file/294a8ed24b1ad22ec2e7efea049b8737-Paper.pdf}.

\bibitem[Arora and Zhang(2017)]{aroraGANsActuallyLearn2017}
S.~Arora and Y.~Zhang.
\newblock Do {{GANs}} actually learn the distribution? {{An}} empirical study.
\newblock \emph{arXiv:1706.08224 [cs]}, 2017.
\newblock \url{http://arxiv.org/abs/1706.08224}.

\bibitem[Batchelor et~al.(2019)Batchelor, Anwaruddin, Ross, Alli, Young, Horne,
  Cestoni, Welt, and Mehran]{batchelorAorticValveStenosis2019}
W.~Batchelor, S.~Anwaruddin, L.~Ross, O.~Alli, M.~N. Young, A.~Horne,
  A.~Cestoni, F.~Welt, and R.~Mehran.
\newblock Aortic {{Valve Stenosis Treatment~Disparities}} in the
  {{Underserved}}: {{JACC Council Perspectives}}.
\newblock \emph{Journal of the American College of Cardiology}, 74\penalty0
  (18), 2019.

\bibitem[Baumgartner et~al.(2017)Baumgartner, Hung, Bermejo, Chambers,
  Edvardsen, Goldstein, Lancellotti, LeFevre, Miller, and
  Otto]{baumgartnerRecommendationsEchocardiographicAssessment2017}
H.~Baumgartner, J.~Hung, J.~Bermejo, J.~B. Chambers, T.~Edvardsen,
  S.~Goldstein, P.~Lancellotti, M.~LeFevre, F.~Miller, and C.~M. Otto.
\newblock Recommendations on the echocardiographic assessment of aortic valve
  stenosis: A focused update from the {{European Association}} of
  {{Cardiovascular Imaging}} and the {{American Society}} of
  {{Echocardiography}}.
\newblock \emph{European Heart Journal Cardiovascular Imaging}, 18\penalty0
  (3):\penalty0 254--275, 2017.

\bibitem[Berthelot et~al.(2019{\natexlab{a}})Berthelot, Carlini, Cubuk,
  Kurakin, Sohn, Zhang, and
  Raffel]{berthelotRemixmatchSemisupervisedLearning2019}
D.~Berthelot, N.~Carlini, E.~D. Cubuk, A.~Kurakin, K.~Sohn, H.~Zhang, and
  C.~Raffel.
\newblock Remixmatch: {{Semi}}-supervised learning with distribution alignment
  and augmentation anchoring.
\newblock \emph{arXiv preprint arXiv:1911.09785}, 2019{\natexlab{a}}.

\bibitem[Berthelot et~al.(2019{\natexlab{b}})Berthelot, Carlini, Goodfellow,
  Papernot, Oliver, and Raffel]{berthelotMixmatchHolisticApproach2019}
D.~Berthelot, N.~Carlini, I.~Goodfellow, N.~Papernot, A.~Oliver, and C.~A.
  Raffel.
\newblock Mixmatch: {{A}} holistic approach to semi-supervised learning.
\newblock In \emph{Advances in Neural Information Processing Systems},
  2019{\natexlab{b}}.

\bibitem[Brennan et~al.(2019)Brennan, Bryant, Boero, Wang, Vemulapalli,
  Thourani, and Peterson]{brennanProviderlevelVariabilityTreatment2019}
M.~J. Brennan, A.~Bryant, I.~Boero, T.~Wang, S.~Vemulapalli, V.~Thourani, and
  E.~Peterson.
\newblock Provider-level variability in the treatment of patients with severe
  symptomatic aortic valve stenosis.
\newblock \emph{Journal of the American College of Cardiology}, 73\penalty0 (9
  (Supplement 1)):\penalty0 1949, 2019.

\bibitem[Calderon-Ramirez et~al.(2021)Calderon-Ramirez, Giri, Yang, Moemeni,
  Umana, Elizondo, Torrents-Barrena, and Molina-Cabello]{calderon2021dealing}
S.~Calderon-Ramirez, R.~Giri, S.~Yang, A.~Moemeni, M.~Umana, D.~Elizondo,
  J.~Torrents-Barrena, and M.~A. Molina-Cabello.
\newblock Dealing with scarce labelled data: Semi-supervised deep learning with
  mix match for covid-19 detection using chest x-ray images.
\newblock In \emph{International Conference on Pattern Recognition (ICPR)},
  2021.

\bibitem[Caruana et~al.(2004)Caruana, {Niculescu-Mizil}, Crew, and
  Ksikes]{caruanaEnsembleSelectionLibraries2004}
R.~Caruana, A.~{Niculescu-Mizil}, G.~Crew, and A.~Ksikes.
\newblock Ensemble selection from libraries of models.
\newblock In \emph{Proceedings of the Twenty-First International Conference on
  {{Machine}} Learning}, 2004.

\bibitem[Chapelle et~al.(2010)Chapelle, Schlkopf, and
  Zien]{chapelleSemiSupervisedLearning2010}
O.~Chapelle, B.~Schlkopf, and A.~Zien.
\newblock \emph{Semi-{{Supervised Learning}}}.
\newblock {The MIT Press}, 1st edition, 2010.

\bibitem[Chen et~al.(2020{\natexlab{a}})Chen, Qin, Qiu, Tarroni, Duan, Bai, and
  Rueckert]{chenDeepLearningCardiac2020}
C.~Chen, C.~Qin, H.~Qiu, G.~Tarroni, J.~Duan, W.~Bai, and D.~Rueckert.
\newblock Deep learning for cardiac image segmentation: {{A}} review.
\newblock \emph{Frontiers in Cardiovascular Medicine}, 7, 2020{\natexlab{a}}.

\bibitem[Chen et~al.(2020{\natexlab{b}})Chen, Kornblith, Swersky, Norouzi, and
  Hinton]{chenBigSelfSupervisedModels2020}
T.~Chen, S.~Kornblith, K.~Swersky, M.~Norouzi, and G.~Hinton.
\newblock Big {{Self}}-{{Supervised Models}} are {{Strong Semi}}-{{Supervised
  Learners}}.
\newblock \emph{arXiv:2006.10029 [cs, stat]}, 2020{\natexlab{b}}.
\newblock \url{http://arxiv.org/abs/2006.10029}.

\bibitem[Chen et~al.(2021)Chen, Wang, Lai, Chen, Cao, Ye, Ren, Zhao, Zhou, and
  Qi]{chen2021venibot}
Y.~Chen, Y.~Wang, B.~Lai, Z.~Chen, X.~Cao, N.~Ye, Z.~Ren, J.~Zhao, X.-Y. Zhou,
  and P.~Qi.
\newblock Venibot: Towards autonomous venipuncture with semi-supervised vein
  segmentation from ultrasound images.
\newblock \emph{arXiv preprint arXiv:2105.12945}, 2021.

\bibitem[Clark et~al.(2012)Clark, Arnold, Duhay, Thompson, Keyes, Svensson,
  Bonow, Stockwell, and Cohen]{clarkFiveyearClinicalEconomic2012}
M.~A. Clark, S.~V. Arnold, F.~G. Duhay, A.~K. Thompson, M.~J. Keyes, L.~G.
  Svensson, R.~O. Bonow, B.~T. Stockwell, and D.~J. Cohen.
\newblock Five-year clinical and economic outcomes among patients with
  medically managed severe aortic stenosis: Results from a {{Medicare}} claims
  analysis.
\newblock \emph{Circulation. Cardiovascular Quality and Outcomes}, 5\penalty0
  (5):\penalty0 697--704, 2012.

\bibitem[Coates et~al.(2011)Coates, Lee, and Ng]{coatesAnalysisSingleLayer2011}
A.~Coates, H.~Lee, and A.~Y. Ng.
\newblock An {{Analysis}} of {{Single Layer Networks}} in {{Unsupervised
  Feature Learning}}.
\newblock In \emph{Artificial {{Intelligence}} and {{Statistics}}}, 2011.
\newblock \url{https://cs.stanford.edu/~acoates/stl10/}.

\bibitem[Ghorbani et~al.(2020)Ghorbani, Ouyang, Abid, He, Chen, Harrington,
  Liang, Ashley, and Zou]{ghorbaniDeepLearningInterpretation2020}
A.~Ghorbani, D.~Ouyang, A.~Abid, B.~He, J.~H. Chen, R.~A. Harrington, D.~H.
  Liang, E.~A. Ashley, and J.~Y. Zou.
\newblock Deep learning interpretation of echocardiograms.
\newblock \emph{npj Digital Medicine}, 3\penalty0 (1), 2020.

\bibitem[Goodfellow et~al.(2014)Goodfellow, {Pouget-Abadie}, Mirza, Xu,
  {Warde-Farley}, Ozair, Courville, and
  Bengio]{goodfellowGenerativeAdversarialNets2014}
I.~Goodfellow, J.~{Pouget-Abadie}, M.~Mirza, B.~Xu, D.~{Warde-Farley},
  S.~Ozair, A.~Courville, and Y.~Bengio.
\newblock Generative {{Adversarial Nets}}.
\newblock In \emph{Advances in {{Neural Information Processing Systems}}},
  2014.
\newblock
  \url{https://papers.neurips.cc/paper/2014/file/5ca3e9b122f61f8f06494c97b1afccf3-Paper.pdf}.

\bibitem[Hata et~al.(2020)Hata, Seo, Nakayama, Iwasaki, Ohkawauchi, and
  Ohya]{hataClassificationAorticStenosis2020}
E.~Hata, C.~Seo, M.~Nakayama, K.~Iwasaki, T.~Ohkawauchi, and J.~Ohya.
\newblock Classification of {{Aortic Stenosis Using ECG}} by {{Deep Learning}}
  and its {{Analysis Using Grad}}-{{CAM}}.
\newblock \emph{Annual International Conference of the IEEE Engineering in
  Medicine and Biology Society. IEEE Engineering in Medicine and Biology
  Society. Annual International Conference}, 2020:\penalty0 1548--1551, 2020.

\bibitem[Howlader et~al.(2020)Howlader, Noone, Krapcho, Miller, Brest, Yu,
  Ruhl, Tatalovich, Mariotto, Lewis, Chen, Feuer, and
  Cronin]{howladerSEERCancerStatistics2020}
N.~Howlader, A.~M. Noone, M.~Krapcho, D.~Miller, A.~Brest, M.~Yu, J.~Ruhl,
  Z.~Tatalovich, A.~Mariotto, D.~R. Lewis, H.~S. Chen, E.~J. Feuer, and K.~A.
  Cronin.
\newblock {{SEER Cancer Statistics Review}}, 1975-2017.
\newblock Technical report, {National Cancer Institute}, 2020.
\newblock \url{https://seer.cancer.gov/csr/1975\\_2017}.

\bibitem[Huang et~al.(2017)Huang, Li, Pleiss, Liu, Hopcroft, and
  Weinberger]{huangSnapshotEnsemblesTrain2017}
G.~Huang, Y.~Li, G.~Pleiss, Z.~Liu, J.~E. Hopcroft, and K.~Q. Weinberger.
\newblock Snapshot ensembles: {{Train}} 1, get m for free.
\newblock \emph{arXiv preprint arXiv:1704.00109}, 2017.

\bibitem[Kingma and Ba(2014)]{kingma2014adam}
D.~P. Kingma and J.~Ba.
\newblock Adam: A method for stochastic optimization.
\newblock \emph{arXiv preprint arXiv:1412.6980}, 2014.

\bibitem[Krizhevsky(2009)]{krizhevskyLearningMultipleLayers2009}
A.~Krizhevsky.
\newblock Learning {{Multiple Layers}} of {{Features}} from {{Tiny Images}}.
\newblock Technical report, {University of Toronto}, 2009.
\newblock \url{https://www.cs.toronto.edu/~kriz/learning-features-2009-TR.pdf}.

\bibitem[Kwon et~al.(2020)Kwon, Lee, Jeon, Lee, Kim, Park, Oh, and
  Lee]{kwonDeepLearningBased2020}
J.-M. Kwon, S.~Y. Lee, K.-H. Jeon, Y.~Lee, K.-H. Kim, J.~Park, B.-H. Oh, and
  M.-M. Lee.
\newblock Deep {{Learning}}\textendash{{Based Algorithm}} for {{Detecting
  Aortic Stenosis Using Electrocardiography}}.
\newblock \emph{Journal of the American Heart Association}, 9\penalty0 (7),
  2020.

\bibitem[Lancellotti et~al.(2018)Lancellotti, Magne, Dulgheru, Clavel, Donal,
  Vannan, Chambers, Rosenhek, Habib, Lloyd, Nistri, Garbi, Marchetta, Fattouch,
  Coisne, Montaigne, Modine, Davin, Gach, Radermecker, Liu, Gillam, Rossi,
  Galli, Ilardi, Tastet, Capoulade, Zilberszac, Vollema, Delgado, Cosyns,
  Lafitte, Bernard, Pierard, Bax, Pibarot, and
  Oury]{lancellottiOutcomesPatientsAsymptomatic2018}
P.~Lancellotti, J.~Magne, R.~Dulgheru, M.-A. Clavel, E.~Donal, M.~A. Vannan,
  J.~Chambers, R.~Rosenhek, G.~Habib, G.~Lloyd, S.~Nistri, M.~Garbi,
  S.~Marchetta, K.~Fattouch, A.~Coisne, D.~Montaigne, T.~Modine, L.~Davin,
  O.~Gach, M.~Radermecker, S.~Liu, L.~Gillam, A.~Rossi, E.~Galli, F.~Ilardi,
  L.~Tastet, R.~Capoulade, R.~Zilberszac, E.~M. Vollema, V.~Delgado, B.~Cosyns,
  S.~Lafitte, A.~Bernard, L.~A. Pierard, J.~J. Bax, P.~Pibarot, and C.~Oury.
\newblock Outcomes of {{Patients With Asymptomatic Aortic Stenosis Followed
  Up}} in {{Heart Valve Clinics}}.
\newblock \emph{JAMA cardiology}, 3\penalty0 (11):\penalty0 1060--1068, 2018.

\bibitem[Leclerc et~al.(2019)Leclerc, Smistad, Pedrosa, {\O}stvik, Cervenansky,
  Espinosa, Espeland, Berg, Jodoin, Grenier, et~al.]{leclerc2019deep}
S.~Leclerc, E.~Smistad, J.~Pedrosa, A.~{\O}stvik, F.~Cervenansky, F.~Espinosa,
  T.~Espeland, E.~A.~R. Berg, P.-M. Jodoin, T.~Grenier, et~al.
\newblock Deep learning for segmentation using an open large-scale dataset in
  2d echocardiography.
\newblock \emph{IEEE transactions on medical imaging}, 38\penalty0 (9), 2019.

\bibitem[Lee(2013)]{leePseudolabelSimpleEfficient2013}
D.-H. Lee.
\newblock Pseudo-label: {{The}} simple and efficient semi-supervised learning
  method for deep neural networks.
\newblock In \emph{Workshop on Challenges in Representation Learning,
  {{ICML}}}, volume~3, 2013.

\bibitem[Long and
  Wessler(2018)]{longIdentificationEchocardiographicImaging2018}
G.~Long and B.~S. Wessler.
\newblock Identification of {{Echocardiographic Imaging View Using Deep
  Learning}} | {{Circulation}}: {{Cardiovascular Quality}} and {{Outcomes}}.
\newblock In \emph{Circulation: {{Cardiovascular Quality}} and {{Outcomes}}},
  2018.
\newblock
  \url{https://www.ahajournals.org/doi/10.1161/circoutcomes.11.suppl_1.276}.

\bibitem[Madani et~al.(2018{\natexlab{a}})Madani, Arnaout, Mofrad, and
  Arnaout]{madaniFastAccurateView2018}
A.~Madani, R.~Arnaout, M.~Mofrad, and R.~Arnaout.
\newblock Fast and accurate view classification of echocardiograms using deep
  learning.
\newblock \emph{npj Digital Medicine}, 1\penalty0 (1), 2018{\natexlab{a}}.

\bibitem[Madani et~al.(2018{\natexlab{b}})Madani, Ong, Tibrewal, and
  Mofrad]{madaniDeepEchocardiographyDataefficient2018}
A.~Madani, J.~R. Ong, A.~Tibrewal, and M.~R. Mofrad.
\newblock Deep echocardiography: Data-efficient supervised and semi-supervised
  deep learning towards automated diagnosis of cardiac disease.
\newblock \emph{NPJ digital medicine}, 1\penalty0 (1), 2018{\natexlab{b}}.

\bibitem[Meng et~al.(2020)Meng, Matthew, Zimmer, Gomez, Lloyd, Rueckert, and
  Kainz]{meng2020mutual}
Q.~Meng, J.~Matthew, V.~A. Zimmer, A.~Gomez, D.~F. Lloyd, D.~Rueckert, and
  B.~Kainz.
\newblock Mutual information-based disentangled neural networks for classifying
  unseen categories in different domains: application to fetal ultrasound
  imaging.
\newblock \emph{IEEE Transactions on Medical Imaging}, 40\penalty0 (2), 2020.

\bibitem[Metz et~al.(2017)Metz, Poole, Pfau, and
  {Sohl-Dickstein}]{metzUnrolledGenerativeAdversarial2017}
L.~Metz, B.~Poole, D.~Pfau, and J.~{Sohl-Dickstein}.
\newblock Unrolled {{Generative Adversarial Networks}}.
\newblock \emph{arXiv:1611.02163 [cs, stat]}, 2017.
\newblock \url{http://arxiv.org/abs/1611.02163}.

\bibitem[Minners et~al.(2008)Minners, Allgeier, {Gohlke-Baerwolf}, Kienzle,
  Neumann, and Jander]{minnersInconsistenciesEchocardiographicCriteria2008}
J.~Minners, M.~Allgeier, C.~{Gohlke-Baerwolf}, R.-P. Kienzle, F.-J. Neumann,
  and N.~Jander.
\newblock Inconsistencies of echocardiographic criteria for the grading of
  aortic valve stenosis.
\newblock \emph{European Heart Journal}, 29\penalty0 (8):\penalty0 1043--1048,
  2008.

\bibitem[Mitchell et~al.(2019)Mitchell, Rahko, Blauwet, Canaday, Finstuen,
  Foster, Horton, Ogunyankin, Palma, and
  Velazquez]{mitchellGuidelinesPerformingComprehensive2019}
C.~Mitchell, P.~S. Rahko, L.~A. Blauwet, B.~Canaday, J.~A. Finstuen, M.~C.
  Foster, K.~Horton, K.~O. Ogunyankin, R.~A. Palma, and E.~J. Velazquez.
\newblock Guidelines for {{Performing}} a {{Comprehensive Transthoracic
  Echocardiographic Examination}} in {{Adults}}: {{Recommendations}} from the
  {{American Society}} of {{Echocardiography}}.
\newblock \emph{Journal of the American Society of Echocardiography: Official
  Publication of the American Society of Echocardiography}, 32\penalty0
  (1):\penalty0 1--64, 2019.

\bibitem[Miyato et~al.(2019)Miyato, Maeda, Koyama, and
  Ishii]{miyatoVirtualAdversarialTraining2019}
T.~Miyato, S.-I. Maeda, M.~Koyama, and S.~Ishii.
\newblock Virtual {{Adversarial Training}}: {{A Regularization Method}} for
  {{Supervised}} and {{Semi}}-{{Supervised Learning}}.
\newblock \emph{IEEE Transactions on Pattern Analysis and Machine
  Intelligence}, 41\penalty0 (8), 2019.
\newblock \url{https://ieeexplore.ieee.org/document/8417973/}.

\bibitem[Netzer et~al.(2011)Netzer, Wang, Coates, Bissacco, Wu, and
  Ng]{netzerReadingDigitsNatural2011}
Y.~Netzer, T.~Wang, A.~Coates, A.~Bissacco, B.~Wu, and A.~Y. Ng.
\newblock Reading {{Digits}} in {{Natural Images}} with {{Unsupervised Feature
  Learning}}.
\newblock In \emph{{{NeurIPS Workshop}} on {{Deep Learning}} and {{Unsupervised
  Feature Learning}}}, 2011.
\newblock \url{http://ufldl.stanford.edu/housenumbers}.

\bibitem[Oliver et~al.(2018)Oliver, Odena, Raffel, Cubuk, and
  Goodfellow]{oliverRealisticEvaluationDeep2018}
A.~Oliver, A.~Odena, C.~A. Raffel, E.~D. Cubuk, and I.~Goodfellow.
\newblock Realistic evaluation of deep semi-supervised learning algorithms.
\newblock In \emph{Advances in Neural Information Processing Systems}, 2018.

\bibitem[Ouyang et~al.(2020)Ouyang, He, Ghorbani, Yuan, Ebinger, Langlotz,
  Heidenreich, Harrington, Liang, Ashley, and
  Zou]{ouyangVideobasedAIBeattobeat2020}
D.~Ouyang, B.~He, A.~Ghorbani, N.~Yuan, J.~Ebinger, C.~P. Langlotz, P.~A.
  Heidenreich, R.~A. Harrington, D.~H. Liang, E.~A. Ashley, and J.~Y. Zou.
\newblock Video-based {{AI}} for beat-to-beat assessment of cardiac function.
\newblock \emph{Nature}, 580\penalty0 (7802), 2020.

\bibitem[Ruder(2017)]{ruder2017overview}
S.~Ruder.
\newblock An overview of multi-task learning in deep neural networks.
\newblock \emph{arXiv preprint arXiv:1706.05098}, 2017.

\bibitem[Sacchi et~al.(2018)Sacchi, Dhutia, {Shun-Shin}, Zolgharni, Sutaria,
  Francis, and Cole]{sacchiDopplerAssessmentAortic2018}
S.~Sacchi, N.~M. Dhutia, M.~J. {Shun-Shin}, M.~Zolgharni, N.~Sutaria, D.~P.
  Francis, and G.~D. Cole.
\newblock Doppler assessment of aortic stenosis: A 25-operator study
  demonstrating why reading the peak velocity is superior to velocity time
  integral.
\newblock \emph{European Heart Journal Cardiovascular Imaging}, 19\penalty0
  (12):\penalty0 1380--1389, 2018.

\bibitem[Selvaraju et~al.(2020)Selvaraju, Cogswell, Das, Vedantam, Parikh, and
  Batra]{selvarajuGradCAMVisualExplanations2020}
R.~R. Selvaraju, M.~Cogswell, A.~Das, R.~Vedantam, D.~Parikh, and D.~Batra.
\newblock Grad-{{CAM}}: {{Visual Explanations}} from {{Deep Networks}} via
  {{Gradient}}-based {{Localization}}.
\newblock \emph{International Journal of Computer Vision}, 128\penalty0
  (2):\penalty0 336--359, 2020.

\bibitem[Shen et~al.(2017)Shen, Wu, and Suk]{shenDeepLearningMedical2017}
D.~Shen, G.~Wu, and H.-I. Suk.
\newblock Deep learning in medical image analysis.
\newblock \emph{Annual Review of Biomedical Engineering}, 19\penalty0
  (1):\penalty0 221--248, 2017.

\bibitem[Simonyan et~al.(2014)Simonyan, Vedaldi, and
  Zisserman]{simonyanDeepConvolutionalNetworks2014}
K.~Simonyan, A.~Vedaldi, and A.~Zisserman.
\newblock Deep {{Inside Convolutional Networks}}: {{Visualising Image
  Classification Models}} and {{Saliency Maps}}.
\newblock In \emph{{{arXiv}}:1312.6034 [Cs]}, 2014.
\newblock \url{http://arxiv.org/abs/1312.6034}.

\bibitem[Sohn et~al.(2020)Sohn, Berthelot, Li, Zhang, Carlini, Cubuk, Kurakin,
  Zhang, and Raffel]{sohnFixmatchSimplifyingSemisupervised2020}
K.~Sohn, D.~Berthelot, C.-L. Li, Z.~Zhang, N.~Carlini, E.~D. Cubuk, A.~Kurakin,
  H.~Zhang, and C.~Raffel.
\newblock Fixmatch: {{Simplifying}} semi-supervised learning with consistency
  and confidence.
\newblock \emph{arXiv preprint arXiv:2001.07685}, 2020.

\bibitem[Tang et~al.(2018)Tang, G{\"o}ssl, Ahmed, Garberich, Bradley, Niikura,
  Witt, Pedersen, Bae, Lesser, Harris, Sun, Mudy, and
  Sorajja]{tangContemporaryReasonsClinical2018}
L.~Tang, M.~G{\"o}ssl, A.~Ahmed, R.~Garberich, S.~M. Bradley, H.~Niikura,
  D.~Witt, W.~R. Pedersen, R.~Bae, J.~R. Lesser, K.~M. Harris, B.~Sun, K.~Mudy,
  and P.~Sorajja.
\newblock Contemporary {{Reasons}} and {{Clinical Outcomes}} for {{Patients
  With Severe}}, {{Symptomatic Aortic Stenosis Not Undergoing Aortic Valve
  Replacement}}.
\newblock \emph{Circulation: Cardiovascular Interventions}, 11\penalty0 (12),
  2018.

\bibitem[{van Engelen} and
  Hoos(2020)]{vanengelenSurveySemisupervisedLearning2020}
J.~E. {van Engelen} and H.~H. Hoos.
\newblock A survey on semi-supervised learning.
\newblock \emph{Machine Learning}, 109\penalty0 (2):\penalty0 373--440, 2020.

\bibitem[Wang et~al.(2021)Wang, Chen, Xiang, Lin, Lin, and Heng]{wang2021deep}
X.~Wang, H.~Chen, H.~Xiang, H.~Lin, X.~Lin, and P.-A. Heng.
\newblock Deep virtual adversarial self-training with consistency
  regularization for semi-supervised medical image classification.
\newblock \emph{Medical image analysis}, 70, 2021.

\bibitem[Xie et~al.(2019)Xie, Dai, Hovy, Luong, and Le]{xie2019unsupervised}
Q.~Xie, Z.~Dai, E.~Hovy, M.-T. Luong, and Q.~V. Le.
\newblock Unsupervised data augmentation for consistency training.
\newblock \emph{arXiv preprint arXiv:1904.12848}, 2019.

\bibitem[Yang et~al.(2020)Yang, Ojha, Aranoff, Green, and
  Tavassolian]{yangClassificationAorticStenosis2020}
C.~Yang, B.~D. Ojha, N.~D. Aranoff, P.~Green, and N.~Tavassolian.
\newblock Classification of aortic stenosis using conventional machine learning
  and deep learning methods based on multi-dimensional cardio-mechanical
  signals.
\newblock \emph{Scientific Reports}, 10\penalty0 (1):\penalty0 17521, 2020.

\bibitem[Zagoruyko and Komodakis(2017)]{zagoruykoWideResidualNetworks2017}
S.~Zagoruyko and N.~Komodakis.
\newblock Wide {{Residual Networks}}.
\newblock \emph{arXiv:1605.07146 [cs]}, 2017.
\newblock \url{http://arxiv.org/abs/1605.07146}.

\bibitem[Zhang et~al.(2017)Zhang, Cisse, Dauphin, and
  {Lopez-Paz}]{zhangMixupEmpiricalRisk2017}
H.~Zhang, M.~Cisse, Y.~N. Dauphin, and D.~{Lopez-Paz}.
\newblock Mixup: {{Beyond}} empirical risk minimization.
\newblock \emph{arXiv preprint arXiv:1710.09412}, 2017.

\bibitem[Zhang et~al.(2018)Zhang, Gajjala, Agrawal, Tison, Hallock,
  {Beussink-Nelson}, Lassen, Fan, Aras, Jordan, Fleischmann, Melisko, Qasim,
  Shah, Bajcsy, and Deo]{zhangFullyAutomatedEchocardiogram2018}
J.~Zhang, S.~Gajjala, P.~Agrawal, G.~H. Tison, L.~A. Hallock,
  L.~{Beussink-Nelson}, M.~H. Lassen, E.~Fan, M.~A. Aras, C.~Jordan, K.~E.
  Fleischmann, M.~Melisko, A.~Qasim, S.~J. Shah, R.~Bajcsy, and R.~C. Deo.
\newblock Fully {{Automated Echocardiogram Interpretation}} in {{Clinical
  Practice}}.
\newblock \emph{Circulation}, 138\penalty0 (16):\penalty0 1623--1635, 2018.

\bibitem[Zhang and Yang(2021)]{zhang2021survey}
Y.~Zhang and Q.~Yang.
\newblock A survey on multi-task learning.
\newblock \emph{IEEE Transactions on Knowledge and Data Engineering}, 2021.

\bibitem[Zhu(2005)]{zhuSemiSupervisedLearningLiterature2005}
X.~Zhu.
\newblock Semi-{{Supervised Learning Literature Survey}}.
\newblock Technical Report Technical Report 1530, {Department of Computer
  Science, University of Wisconsin Madison.}, 2005.

\end{thebibliography}
